\documentclass{new_tlp}

\usepackage[utf8]{inputenc}
\usepackage{amsmath}
\usepackage{amssymb}
\usepackage{microtype}
\usepackage{url}
\usepackage{hyperref}

\newcommand{\tuple}[1]{\langle #1 \rangle}
\renewcommand{\H}{\ensuremath{\mathbf{H}}} 
\newcommand{\T}{\ensuremath{\mathbf{T}}}
\newcommand{\M}{\ensuremath{\mathbf{M}}}

\input{include/asp-macros/dtel}
\newcommand{\HT}{\ensuremath{\mathrm{HT}}}

\newcommand{\LTL}{\ensuremath{\mathrm{LTL}}}

\newcommand{\THT}{\ensuremath{\mathrm{THT}}}

\newcommand{\TEL}{\ensuremath{\mathrm{TEL}}}

\newcommand{\DEL}{\ensuremath{\mathrm{DEL}}}

\newcommand{\DL}{\ensuremath{\mathrm{DL}}}

\newcommand{\MTL}{\ensuremath{\mathrm{MTL}}}
\newcommand{\MTLf}{\ensuremath{\mathrm{MTL}_{\!f}}}

\newcommand{\MHT}{\ensuremath{\mathrm{MHT}}}
\newcommand{\MHTf}{\ensuremath{\mathrm{MHT}_{\!f}}}
\newcommand{\MHTo}{\ensuremath{\mathrm{MHT}_{\!\omega}}}
\newcommand{\MEL}{\ensuremath{\mathrm{MEL}}}
\newcommand{\MELf}{\ensuremath{{\MEL}_{\!f}}}
\newcommand{\MELo}{\ensuremath{{\MEL}_{\omega}}}

\newcommand{\sysfont}{\textit}

\newcommand{\clingo}{\sysfont{clingo}}

\newcommand{\telingo}{\sysfont{telingo}}

\usepackage{ifthen}
\newcommand{\eqdef}{\ensuremath{\mathbin{\raisebox{-1pt}[-3pt][0pt]{$\stackrel{\mathit{def}}{=}$}}}}

\newcommand{\myatom}{\ensuremath{p}}
\newcommand{\myatomb}{\ensuremath{q}}
\newcommand{\PV}{\ensuremath{\mathcal{A}}}

\newcommand{\gequiv}{\ensuremath{\mathrel{\equiv}}}

\def\cI{I}
\def\cJ{J}

\newcommand{\metricI}[1]{\ensuremath{#1_{\cI}}}
\newcommand{\metric}[3]{\ensuremath{#1_{
    \ifthenelse{\equal{#2}{#3}}
    {#2}
    {
      \ifthenelse{\equal{#2}{0}}
      {
        \ifthenelse{\equal{#3}{\omega}}
        {}
        {\leq#3}
      }
      {
        \ifthenelse{\equal{#3}{\omega}}
        {\geq#2}
        {\intervc{#2}{#3}}
      }
    }}}}

\newcommand{\tmf}{\ensuremath{\tau}}

\newcommand{\tr}[1]{[#1]}

\newcommand{\QHT}{\ensuremath{\mathit{QHT}}} 
\def\dd{\delta}
\newcommand{\QHTD}{\ensuremath{\mathit{QHT}[\peq{\dd}]}} 
\newcommand{\QHTS}{\ensuremath{\mathit{QHT}^s}} 
\newcommand{\peq}[1]{\preccurlyeq_{#1}}
\newcommand{\pt}[1]{\prec_{#1}}
\newcommand{\metricMN}[1]{\ensuremath{#1_{\intervco{m}{n}}}}
\newcommand{\Atoms}[1]{\ensuremath{\mathit{Atoms}(#1)}}
\def\cP{\mathcal{P}}

\newcommand{\qed}{\ignorespaces} 
\newtheorem{definition}{Definition}
\newtheorem{theorem}{Theorem}
\newtheorem{proposition}{Proposition}
\newtheorem{corollary}{Corollary}
\newtheorem{lemma}{Lemma}
\newenvironment{proofof}[1]{\noindent {\bf Proof of #1.}}{\bigskip}

\newcommand{\EM}[1]{\ensuremath{\mathrm{EM}(#1)}}

\newcommand{\metricol}[3]{\ensuremath{#1_{
    \ifthenelse{\equal{#2}{#3}}
    {#2}
    {
      \ifthenelse{\equal{#2}{0}}
      {
        \ifthenelse{\equal{#3}{\omega}}
        {}
        {\leq#3}
      }
      {
        \ifthenelse{\equal{#3}{\omega}}
        {\geq#2}
        {\intervo{#2}{#3}}
      }
    }}}}

\newcommand{\intervoo}[2]{\ensuremath{(#1..#2)}}
\newcommand{\rangeoo}[3]{\ensuremath{#1 \in \intervoo{#2}{#3}}}
\newcommand{\intervoc}[2]{\ensuremath{(#1..#2]}}
\newcommand{\rangeoc}[3]{\ensuremath{#1 \in \intervoc{#2}{#3}}}

\newcommand{\intervco}[2]{\ensuremath{[#1..#2)}}
\newcommand{\rangeco}[3]{\ensuremath{#1 \in \intervco{#2}{#3}}}
\newcommand{\intervcc}[2]{\ensuremath{[#1..#2]}}
\newcommand{\rangecc}[3]{\ensuremath{#1 \in \intervcc{#2}{#3}}}
\newtheorem{observation}{Observation}

\usepackage{comments}

\begin{document}

\title[Metric Equilibrium Logic]{Metric Temporal Equilibrium Logic\\ over Timed Traces\thanks{A prior version of this paper appeared at LPNMR'22.}}%

\author[Becker et al.]{%
  ARVID BECKER\\                   {University of Potsdam, Germany}
  \and
  PEDRO CABALAR\\                  {University of Corunna, Spain}
  \and
  MART\'{I}N DI\'EGUEZ\\           {LERIA, Universit\'e d'Angers, France}
  \and
  TORSTEN SCHAUB, ANNA SCHUHMANN\\ {University of Potsdam, Germany}
}

\maketitle

\begin{abstract}
  In temporal extensions of Answer Set Programming (ASP) based on linear-time,
  the behavior of dynamic systems is captured by sequences of states.
  While this representation reflects their relative order,
  it abstracts away the specific times associated with each state.
  However, timing constraints are important in many applications like, for instance, when planning and scheduling go hand in hand.
  We address this by developing a metric extension of linear-time temporal equilibrium logic,
  in which temporal operators are constrained by intervals over natural numbers.
  The resulting Metric Equilibrium Logic provides the foundation of an ASP-based approach for specifying
  qualitative and quantitative dynamic constraints.
  To this end,
  we define a translation of metric formulas into monadic first-order formulas and
  give a correspondence between their models in Metric Equilibrium Logic and Monadic Quantified Equilibrium Logic, respectively.
  Interestingly, our translation provides a blue print for implementation in terms of ASP modulo difference constraints.
\end{abstract}
%

\section{Introduction}\label{sec:introduction}

Reasoning about actions and change, or more generally about dynamic systems,
is not only central to knowledge representation and reasoning
but at the heart of Computer Science~\cite{TIMEHandbook}.
In practice, this kind of reasoning often requires both qualitative as well as quantitative dynamic constraints.
For instance, when planning and scheduling at once,
actions may have durations and their effects may need to meet deadlines.
On the other hand, any flexible formalism for actions and change must incorporate some form of non-monotonic reasoning to deal with inertia and other types of defaults.

Over the last years, we addressed qualitative dynamic constraints by combining traditional approaches,
like Dynamic and Linear Temporal Logic (\DL~\cite{hatiko00a} and \LTL~\cite{pnueli77a}),
with the base logic of Answer Set Programming (ASP~\cite{lifschitz99b}), namely,
the logic of Here-and-There (\HT~\cite{heyting30a}) and its non-monotonic extension, called Equilibrium Logic~\cite{pearce96a}. 
This resulted in non-monotonic linear dynamic and temporal equilibrium logics
(\DEL~\cite{bocadisc18a,cadisc19a} and \TEL~\cite{agcadipevi13a,cakascsc18a,agcadipescscvi023})
that gave rise to the temporal ASP system \telingo~\cite{cakamosc19a,cadilasc20a} extending the ASP system
\clingo~\cite{gekakaosscwa16a}.

A commonality of such dynamic and temporal logics is that they abstract from specific time points when capturing temporal relationships.
For instance, in temporal logic, we can use the formula
\(
\alwaysF ( \mathit{use} \to \eventuallyF \mathit{clean})
\)
to express that a machine has to be eventually cleaned after being used.
Nothing can be said about the delay between using and cleaning the machine.

A key design decision was to base both aforementioned logics, \TEL\ and \DEL, on the same linear-time semantics.
%
We continued to maintain the same linear-time semantics,
embodied by sequences of states,
when elaborating upon a first ``light-weight'' metric temporal extension of \HT~\cite{cadiscsc20a}.
The ``light-weightiness'' is due to treating time as a state counter by identifying the next time with the next state.
For instance, this allows us to refine our example by stating that, if the machine is used,
it has to be cleaned within the next 3 states, viz.\
\(
\alwaysF (\mathit{use} \to \metric{\eventuallyF}{1}{3}\mathit{clean})
\).
Although this permits the restriction of temporal operators to subsequences of states,
no fine-grained timing constraints are expressible.
In other words, it is as if state transitions were identified with time clicks, and the two things could not be dissociated.

In this paper,
we overcome this limitation by dealing with \emph{timed traces} where each state has an associated \emph{time}, as done in Metric Temporal Logic (\MTL~\cite{koymans90a}).
This allows us to measure time differences between events.
For instance, in our example, we may thus express that whenever the machine is used, it has to be cleaned within 60 to 120 time units, by writing:
\begin{align*}\label{ex:use:clean}
  \alwaysF (\mathit{use} \to \metric{\eventuallyF}{60}{120}\mathit{clean}) \ .
\end{align*}
Unlike the non-metric version,
this stipulates that
once $\mathit{use}$ is true in a state,
$\mathit{clean}$ must be true in some future state
whose associated time is at least 60 and at most 120 time units after the time of $\mathit{use}$.
The choice of time domain is crucial, and might even lead to undecidability in the continuous case.
We rather adapt a discrete approach that offers a sequence of snapshots of a dynamic system.

The definition of the new variant of \emph{Metric (Temporal) Equilibrium Logic} (\MEL\ for short) is done in two steps.
We start with the definition of a monotonic logic called \emph{Metric (Temporal) logic of Here-and-There} (\MHT),
a temporal extension of the intermediate logic of Here-and-There, mentioned above.
We then select some models from \MHT\ that are said to be in equilibrium, obtaining in this way a non-monotonic entailment relation.

The rest of the paper is organized as follows.
In the next section, we start describing the monotonic basis, \MHT, that generalizes~\cite{cadiscsc20a} by adding timed traces, and provide some basic properties and useful equivalences in this logic.
In Section~\ref{sec:mel}, we study the non-monotonic formalism, \MEL, providing the definition of metric equilibrium models as a kind of minimal \MHT\ models.
We also illustrate this definition with an example and discuss the property of strong equivalence for metric theories, proving that it coincides with equivalence in the monotonic logic, \MHT.
Section~\ref{sec:kamp} provides a translation of \MHT\ into a fragment of first-order \HT,
called \emph{Quantified Here-and-There with Difference Constraints},
following a similar spirit to the well-known translation of Kamp~\citeyear{kamp68a} from \LTL\ to first-order logic.
Finally, Section~\ref{sec:discussion} contains a discussion and concludes the paper.
\ref{sec:appendix} includes all the proofs of the results in the paper.


\section{Metric Logic of Here-and-There}\label{sec:approach}

In this section, we start describing the metric extension of \HT, called \MHT, that is used as monotonic basis for defining Metric Equilibrium Logic later on.
We begin introducing some notation.
Given $m \in \mathbb{N}$ and $n \in \mathbb{N} \cup \{\omega\}$, we let
\intervc{m}{n} stand for the set $\{i \in \mathbb{N} \mid m \leq i \leq n\}$,
\intervo{m}{n}       for         $\{i \in \mathbb{N} \mid m \leq i < n\}$, and
\ointerv{m}{n} stand for $\{i \in \mathbb{N} \mid m < i \leq n\}$.
We use letters $\cI, \cJ$ to denote intervals and, since they stand for sets, we assume standard set relations like inclusion $\cI \subseteq \cJ$ or membership $n \in I$.

Given a set \PV\ of propositional variables (called \emph{alphabet}),
a \emph{metric formula} $\varphi$ is defined by the grammar:
\[
\varphi ::= \myatom \mid \bot \mid \varphi_1 \otimes \varphi_2 \mid
\metricI{\previous}\varphi \mid
\varphi_1 \metricI{\since} \varphi_2 \mid
\varphi_1 \metricI{\trigger} \varphi_2 \mid
\metricI{\next} \varphi \mid
\varphi_1 \metricI{\until} \varphi_2
\mid \varphi_1 \metricI{\release} \varphi_2
\]
where $\myatom \in\PV$ is an atom and $\otimes$ is any binary Boolean connective $\otimes \in \{\to,\wedge,\vee\}$.
The last six cases above correspond to temporal operators,
each of them indexed by some interval $\cI$ of the form \intervo{m}{n} with $m\in\mathbb{N}$ and $n\in\mathbb{N}\cup\{\omega\}$.
In words,
\metricI{\previous},
\metricI{\since}, and
\metricI{\trigger}
are past operators called
\emph{previous},
\emph{since}, and
\emph{trigger}, respectively;
their future counterparts
\metricI{\next},
\metricI{\until}, and
\metricI{\release}
are called
\emph{next},
\emph{until}, and
\emph{release}.
Strictly speaking, we should differentiate between the syntactic representation of an interval, and its semantic counterpart, the associated set of time points it represents.
For simplicity, we just use the same representation for both concepts but, as said above, we restrict the form of intervals that can be used as modal subindices to the case $\intervo{m}{n}$ where $n$ is possibly $\omega$.
Yet, some syntactic abbreviations are allowed in the temporal subindices.
For instance, we let subindex \intervc{m}{n} stand for \intervo{m}{n{+}1}, provided $n\neq\omega$.
Also,
we sometimes use the subindices `${\leq}n$', `${\geq} m$' and `$m$' as abbreviations of intervals $\intervc{0}{n}$, $\intervo{m}{\omega}$ and $\intervc{m}{m}$, respectively.
Also, whenever $\cI=\intervo{0}{\omega}$, we simply omit subindex $\cI$.

We also define several common derived operators like the Boolean connectives
\(
\top \eqdef \neg \bot
\),
\(
\neg \varphi \eqdef  \varphi \to \bot
\),
\(
\varphi \leftrightarrow \psi \eqdef (\varphi \to \psi) \wedge (\psi \to \varphi)
\),
and the following temporal operators:
\[
\begin{array}{rcll}
       \metricI{\alwaysP} \varphi  & \eqdef & \bot \metricI{\trigger} \varphi                                & \text{\emph{always before}} \\
   \metricI{\eventuallyP} \varphi  & \eqdef & \top \metricI{\since} \varphi                                  & \text{\emph{eventually before}} \\
                       \initially\,& \eqdef & \neg \metric{\previous}{0}{\omega} \top                        & \text{\emph{initial}}\\
     \metricI{\wprevious} \varphi  & \eqdef & \metricI{\previous} \top \to \metricI{\previous} \varphi& \text{\emph{weak previous}}
\end{array}
\quad
\begin{array}{rcll}
       \metricI{\alwaysF} \varphi  & \eqdef & \bot \metricI{\release} \varphi                                & \text{\emph{always afterward}}\\
   \metricI{\eventuallyF} \varphi  & \eqdef & \top \metricI{\until} \varphi                                  & \text{\emph{eventually afterward}}\\
                         \finally  & \eqdef & \neg \metric{\next}{0}{\omega} \top                            & \text{\emph{final}}\\
         \metricI{\wnext} \varphi  & \eqdef &  \metricI{\next} \top \to \metricI{\next} \varphi        & \text{\emph{weak next}}
\end{array}
\]
Note that \emph{initial} and \emph{final} are not indexed by any interval;
they only depend on the state of the trace,
not on the actual time associated with this state.
On the other hand, the weak version of \emph{next} can no longer be defined in terms of \emph{final}, as done in~\cite{cakascsc18a} with non-metric $\wnext \varphi \equiv \next \varphi \vee \finally$.
For the metric case $\metricI{\wnext} \varphi$, the implication $\metricI{\next} \top \to \metricI{\next} \varphi$
(or equivalently the disjunction $\metricI{\next} \varphi \vee \neg \metricI{\next} \top$) must be used instead,
in order to keep the usual dualities among operators (the same applies to weak \emph{previous}).

A \emph{metric theory} is a (possibly infinite) set of metric formulas.
As an example of a metric theory, we may consider the following scenario for modeling the behavior of traffic lights.
While the light is red by default,
it changes to green within less than 15 time units (say, seconds) whenever the button is pushed;
and it stays green for another 30 seconds at least.
This can be represented as follows.
\begin{align}
\alwaysF ( \mathit{red} \wedge \mathit{green} \to \bot) \label{ex:traffic:light}\\
\alwaysF ( \neg\mathit{green} \to \mathit{red} ) \label{ex:traffic:light:default}\\
\alwaysF \big( \mathit{push} \to {\eventuallyF}_{\intervo{1}{15}}(\metric{\alwaysF}{0}{30}\;  \mathit{green}) \big) \label{ex:traffic:light:push}
\end{align}
Note that this example combines a default rule \eqref{ex:traffic:light:default} with a metric rule \eqref{ex:traffic:light:push},
describing the initiation and duration period of events.
This nicely illustrates the interest in non-monotonic metric representation and reasoning methods.

A \emph{Here-and-There trace} (for short \emph{\HT-trace}) of length $\lambda \in \mathbb{N} \cup \{\omega\}$ over alphabet \PV\ is a sequence of pairs
\(
(\tuple{H_i,T_i})_{\rangeo{i}{0}{\lambda}}
\)
with $H_i\subseteq T_i\subseteq\PV$ for any $\rangeo{i}{0}{\lambda}$.
For convenience, we usually represent an \HT-trace as the pair $\tuple{\H,\T}$ of sequences
$\H = (H_i)_{\rangeo{i}{0}{\lambda}}$ and $\T = (T_i)_{\rangeo{i}{0}{\lambda}}$.
Notice that, when $\lambda=\omega$, this covers traces of infinite length.
We say that $\tuple{\H,\T}$ is \emph{total} whenever $\H=\T$, that is, $H_i=T_i$ for all $\rangeo{i}{0}{\lambda}$.
%

\begin{definition}\label{def:timed:trace}
  A \emph{timed} \HT-trace $(\tuple{\H,\T},\tmf)$ of length $\lambda$ over $(\mathbb{N},<)$ and alphabet $\mathcal{A}$ is a pair consisting of
  \begin{itemize}
  \item an \HT-trace $\tuple{\H,\T} $ of length $\lambda$ over $\mathcal{A}$ and
  \item a function $\tmf: \intervo{0}{\lambda} \to \mathbb{N}$
    such that $\tmf(i)\leq \tmf(i{+}1)$.
  \end{itemize}
  A timed \HT-trace of length $\lambda > 1$ is called \emph{strict} if $\tmf(i)<\tmf(i{+}1)$
  for all $i{+}1 \in \intervoo{0}{\lambda}$ such that $i+1 < \lambda$ and \emph{non-strict} otherwise.
  We assume w.l.o.g.\ that $\tmf(0)=0$. \qed
\end{definition}
%
Function \tmf\ assigns to each state index $i \in \intervo{0}{\lambda}$ a time point $\tmf(i) \in \mathbb{N}$
representing the number of time units (seconds, milliseconds, etc, depending on the chosen granularity)
elapsed since time point $\tmf(0)=0$, chosen as the beginning of the trace.
The difference to the variant of \MHT\ presented in \cite{cadiscsc20a} boils down to the choice of function \tmf.
Essentially, the latter corresponds now to the case where $\tmf$ is the identity function on the interval $\intervo{0}{\lambda}$.

Given any timed \HT-trace,
satisfaction of formulas is defined as follows.
%
\begin{definition}[\MHT-satisfaction]\label{def:mht:satisfaction}%
A timed \HT-trace $\M=(\tuple{\H,\T}, \tmf)$
of length $\lambda$ over alphabet \PV\
  \emph{satisfies} a metric formula $\varphi$ at step $\rangeo{k}{0}{\lambda}$,
  written \mbox{$\M,k \models \varphi$}, if the following conditions hold:
  \begin{enumerate}
  \item $\M,k \not\models \bot$
  \item $\M,k \models \myatom$ if $\myatom \in H_k$ for any atom $\myatom \in \PV$
  \item \label{def:mhtsat:and} $\M, k \models \varphi \wedge \psi$
    iff
    $\M, k \models \varphi$
    and
    $\M, k \models \psi$
  \item \label{def:mhtsat:or} $\M, k \models \varphi \vee \psi$
    iff
    $\M, k \models \varphi$
    or
    $\M, k \models \psi$
  \item $\M, k \models \varphi \to \psi$
    iff
    $\M', k \not \models \varphi$
    or
    $\M', k \models  \psi$, for both $\M'=\M$ and $\M'=(\tuple{\T,\T}, \tmf)$
  \item \label{def:mhtsat:previous} $\M, k \models \metricI{\previous}\, \varphi$
    iff
    $k>0$ and $\M, k{-}1 \models \varphi$ and $\tmf(k)-\tmf(k{-}1) \in \cI$
  \item \label{def:mhtsat:since}$\M, k \models \varphi \metricI{\since} \psi$
    iff
    for some $\rangec{j}{0}{k}$
    with
    $\tmf(k)-\tmf(j) \in \cI$,
    we have
    $\M, j \models \psi$
    and
    $\M, i \models \varphi$ for all $\orange{i}{j}{k}$
  \item \label{def:mhtsat:trigger}$\M, k \models \varphi \metricI{\trigger} \psi$
    iff
    for all $\rangec{j}{0}{k}$
    with
    $\tmf(k)-\tmf(j) \in \cI$,
    we have
    $\M, j \models \psi$
    or
    $\M, i \models \varphi$ for some $\orange{i}{j}{k}$
  \item \label{def:mhtsat:next}$\M, k \models \metricI{\next}\, \varphi$
    iff
    $k+1<\lambda$ and $\M, k{+}1 \models \varphi$
    and $\tmf(k{+}1)-\tmf(k) \in \cI$
  \item \label{def:mhtsat:until}$\M, k \models \varphi \metricI{\until} \psi$
    iff
    for some $\rangeo{j}{k}{\lambda}$
    with
    $\tmf(j)-\tmf(k) \in \cI$,
    we have
    $\M, j \models \psi$
    and
    $\M, i \models \varphi$ for all $\rangeo{i}{k}{j}$
  \item \label{def:mhtsat:release} $\M, k \models \varphi \metricI{\release} \psi$
    iff
    for all $\rangeo{j}{k}{\lambda}$
    with
    $\tmf(j)-\tmf(k) \in \cI$,
    we have
    $\M, j \models \psi$
    or
    $\M, i \models \varphi$ for some $\rangeo{i}{k}{j}$
    \qed
  \end{enumerate}
\end{definition}
%
Satisfaction of derived operators can be easily deduced:
%
\begin{proposition}\label{prop:satisfaction:tel}
  Let $\M=(\tuple{\H,\T},\tmf)$
  be a timed \HT-trace of length $\lambda$ over \PV.
  Given the respective definitions of derived operators, we get the following satisfaction conditions:
  \begin{enumerate} \setcounter{enumi}{11}
  \item $\M, k \models \initially$
    iff
    $k =0$
  \item $\M, k \models \metricI{\wprevious}\, \varphi$
    iff
    $k =0$ or
    $\M, k{-}1 \models \varphi$
    or $\tmf(k)-\tmf(k{-}1) \not\in \cI$
  \item $\M, k \models \metricI{\eventuallyP}\, \varphi$
    iff
    $\M, i \models \varphi$ for some $\rangec{i}{0}{k}$
    with
    $\tmf(k)-\tmf(i) \in \cI$
  \item $\M, k \models \metricI{\alwaysP}\, \varphi$
    iff
    $\M, i \models \varphi$ for all $\rangec{i}{0}{k}$
    with
    $\tmf(k)-\tmf(i) \in \cI$
  \item $\M, k \models \finally$
    iff
    $k+1 = \lambda$
  \item \label{def:mhtsat:wnext} $\M, k \models \metricI{\wnext}\, \varphi$
    iff
    $k+1 = \lambda$
    or $\M, k{+}1 \models \varphi$
    or $\tmf(k{+}1)-\tmf(k) \not\in \cI$
  \item \label{def:mhtsat:eventuallyF} $\M, k \models \metricI{\eventuallyF}\, \varphi$
    iff
    $\M, i \models \varphi$ for some $\rangeo{i}{k}{\lambda}$
    with
    $\tmf(i)-\tmf(k) \in \cI$
  \item \label{def:mhtsat:alwaysF} $\M, k \models \metricI{\alwaysF}\, \varphi$
    iff
    $\M, i \models \varphi$ for all $\rangeo{i}{k}{\lambda}$
    with
    $\tmf(i)-\tmf(k) \in \cI$
  \qed
  \end{enumerate}
\end{proposition}

A formula $\varphi$ is a \emph{tautology} (or is valid), written $\models \varphi$, iff
$\M,k \models \varphi$ for any timed \HT-trace $\M$ and any $k \in \intervo{0}{\lambda}$.
\MHT\ is the logic induced by the set of all such tautologies.
For two formulas $\varphi, \psi$ we write $\varphi \equiv \psi$, iff
$\models \varphi \leftrightarrow \psi$, that is,
$\M,k \models \varphi \leftrightarrow \psi$ for any
timed \HT-trace \M\ of length $\lambda$ and any $k \in \intervo{0}{\lambda}$.
A timed \HT-trace $\M$ is an \MHT\ \emph{model} of a metric theory $\Gamma$ if $\M,0 \models \varphi$ for all $\varphi \in \Gamma$.
The set of \MHT\ models of $\Gamma$ having length $\lambda$ is denoted as $\MHT(\Gamma,\lambda)$,
whereas $\MHT(\Gamma) \eqdef \bigcup_{\lambda=0}^\omega \MHT(\Gamma,\lambda)$ is
the set of all \MHT\ models of $\Gamma$ of any length.
We may obtain fragments of any metric logic by imposing restrictions on the timed traces used for defining tautologies and models.
%
That is, \MHTf\ stands for the restriction of \MHT\ to traces of any finite length $\lambda \in \mathbb{N}$ and
\MHTo\ corresponds to the restriction to traces of infinite length $\lambda=\omega$.

We say that a metric theory is \emph{temporal} if all its modal operators are subindex-free.
Temporal formulas share the same syntax as \LTL,
although the absence of an interval in \MHT\ is understood as an abbreviation for the fixed interval $\intervo{0}{\omega}$.
The following result shows that, for temporal theories, \MHT\ satisfaction collapses into Temporal Here-and-There (\THT)
satisfaction~\cite{agcadipescscvi023}.
Hence, we can use non-timed traces and ignore function \tmf\ in this case.
%
\begin{proposition}\label{prop:tht-fragment}
Let $\Gamma$ be a temporal theory.
Then $(\tuple{\H, \T}, \tmf) \models \Gamma$ in \MHT\ iff $\tuple{\H, \T} \models \Gamma$ in \THT.
\end{proposition}

An interesting subset of \MHT\ is the one formed by total timed traces like $(\tuple{\T, \T}, \tmf)$.
In the non-metric version of temporal \HT, the restriction to
total models corresponds to Linear Temporal Logic (\LTL~\cite{pnueli77a}).
In our case, the restriction to total traces defines a metric version of \LTL,
which we call \emph{Metric Temporal Logic} (or \MTL\ for short).
We present next several properties about total traces and the relation between \MHT\ and \MTL.
%
\begin{proposition}[Persistence]\label{prop:persistence}
  Let $(\tuple{\H,\T},\tmf)$
  be a timed \HT-trace of length $\lambda$ over \PV\ and let $\varphi$ be a metric formula over \PV.
  Then, for any $\rangeo{k}{0}{\lambda}$,
  if $(\tuple{\H,\T}, \tmf), k \models \varphi$ then $(\tuple{\T,\T}, \tmf), k \models \varphi$.\qed
\end{proposition}
%
Thanks to Proposition~\ref{prop:persistence} and a decidability result in~\cite{ouawor07a}, we get:
%
\begin{corollary}[Decidability of \MHTf]\label{cor:decidability}
The logic of \MHTf\ is decidable.\qed
\end{corollary}
%
The next result shows that the satisfaction of negated formulas as classical ones also extends from \HT\ to \MHT:
%
\begin{proposition}\label{prop:negation}
  Let $(\tuple{\H,\T},\tmf)$
  be a timed \HT-trace of length $\lambda$ over \PV\ and let $\varphi$ be a metric formula over \PV.
  Then, $(\tuple{\H,\T}, \tmf), k \models \neg \varphi$ iff $(\tuple{\T,\T}, \tmf), k \not\models \varphi$.\qed
\end{proposition}

In the non-metric case, \LTL\ models can be obtained from \THT\ by adding a particular axiom schema, we call the
\emph{temporal excluded middle} axiom.
%
\begin{definition}[Temporal Excluded Middle]
Given a set of propositional variables \PV, we define the theory \EM{\PV} as
\begin{align*}
  \EM{\PV}& \eqdef \lbrace \alwaysF(p \vee \neg p) \mid p \in \PV \rbrace.
\end{align*}
\end{definition}
%
This same axiom schema can also be used to reduce \MHT\ to \MTL, assuming that, in our current context, operator
$\alwaysF$ stands for $\alwaysF_{\intervo{0}{\omega}}$ as detailed above.
%
\begin{proposition}\label{prop:total}
  Let $\PV$ be a set of atoms.
  For all \MHT\ interpretation $(\tuple{\H,\T},\tmf)$ over \PV,  we have that $(\tuple{\H,\T},\tmf), 0 \models \EM{\PV}$ iff $\H=\T$.
\end{proposition}
%
\begin{corollary}
Let $\Gamma$ be a metric theory over alphabet $\PV$.
The \MTL\ models of $\Gamma$ coincide with the \MHT\ models of $\Gamma \cup \EM{\PV}$.
\end{corollary}

Interestingly, if an equivalence does not involve implication (or negation), we can just check it by only considering total models:
%
\begin{proposition}\label{prop:nonimpl}
  Let $\varphi$ and $\psi$ be metric formulas without implication (and so, without negation either).
  Then, $\varphi \equiv \psi$ in \MTL\ iff $\varphi \equiv \psi$ in \MHT.\qed
\end{proposition}

Many tautologies in \MHT\ or its fragments have a dual version depending on the nature of the operators involved.
This can be formally exploited to save proof effort.
%
We define all pairs of dual connectives as follows:
$\top/\bot$,
$\wedge/\vee$,
$\metricI{\until}/\metricI{\release}$,
$\metricI{\next}/\metricI{\wnext}$,
$\metricI{\alwaysF}/\metricI{\eventuallyF}$,
$\metricI{\since}/\metricI{\trigger}$,
$\metricI{\previous}/\metricI{\wprevious}$,
$\metricI{\alwaysP}/\metricI{\eventuallyP}$.
For any formula $\varphi$ without implications, we define $\delta(\varphi)$
as the result of replacing each connective by its dual operator.

Then, we get the following corollary of Proposition~\ref{prop:nonimpl}.
%
\begin{corollary}[Boolean Duality]\label{BDT}
  Let $\varphi$ and $\psi$ be metric formulas without implication.
  Then, we have in \MHT\ that $\varphi \gequiv \psi$  iff $\delta(\varphi) \gequiv \delta(\psi)$.\qed
\end{corollary}

Let
$\metricI{\until}/\metricI{\since}$,
$\metricI{\release}/\metricI{\trigger}$,
$\metricI{\next}/\metricI{\previous}$,
$\metricI{\wnext}/\metricI{\wprevious}$,
$\metricI{\alwaysF}/\metricI{\alwaysP}$, and
$\metricI{\eventuallyF}/\metricI{\eventuallyP}$
be all pairs of swapped-time connectives and
$\sigma(\varphi)$ be the replacement in $\varphi$ of each connective by its swapped-time version.
Then, we have the following result for finite traces.
%

\begin{lemma}\label{TDT}
  Let \M\ be a finite timed \HT-trace of length $\lambda \in \mathbb{N}$.
  There exists a mapping $\varrho$ from \M\ to another \HT-trace $\varrho(\M)$ of the same length
  such that for any $\rangeo{k}{0}{\lambda}$,
  $\M,k \models \varphi$ iff $\varrho(\M),\lambda\!-\!1\!-\!k \models \sigma(\varphi)$.
\end{lemma}
%
\begin{theorem}[Temporal Duality Theorem]\label{th:temporal_duality}
  A metric formula $\varphi$ is a \MHTf-tautology iff $\sigma(\varphi)$ is a \MHTf-tautology.\qed
\end{theorem}

The next properties capture some distributivity laws for temporal operators with respect to conjunction and disjunction.
%
\begin{proposition}\label{prop:equivalences1}
  For all metric formulas $\varphi$, $\psi$, and $\chi$, the following equivalences hold in \MHT:
  \begin{align*}
    \metricI{\next} \left( \varphi \vee \psi \right) &\equiv \metricI{\next} \varphi \vee \metricI{\next} \psi & \varphi \metricI{\until} \left(\chi \vee \psi\right) &\equiv \left(\varphi \metricI{\until} \chi\right) \vee \left(\varphi \metricI{\until} \psi\right) \\
    \metricI{\next} \left( \varphi \wedge \psi \right) &\equiv \metricI{\next} \varphi \wedge \metricI{\next} \psi &  \left(\varphi\wedge \chi \right) \metricI{\until} \psi &\equiv \left(\varphi \metricI{\until} \psi\right) \wedge \left(\chi \metricI{\until} \psi\right)\\
    \metricI{\wnext} \left( \varphi \vee \psi \right) &\equiv \metricI{\wnext} \varphi \vee \metricI{\wnext} \psi & \varphi \metricI{\release} \left(\chi \wedge \psi\right) &\equiv \left(\varphi \metricI{\release} \chi\right) \wedge \left(\varphi \metricI{\release} \psi\right)\\
    \metricI{\wnext} \left( \varphi \wedge \psi \right) &\equiv \metricI{\wnext} \varphi \wedge \metricI{\wnext} \psi & \left(\varphi\vee \chi \right) \metricI{\release} \psi &\equiv \left(\varphi \metricI{\release} \psi\right) \vee \left(\chi \metricI{\release} \psi\right)\\
    \metricI{\eventuallyF} \left( \varphi \vee \psi \right) &\equiv \metricI{\eventuallyF} \varphi \vee \metricI{\eventuallyF} \psi & \left(\varphi\wedge \chi \right) \metricI{\since} \psi  & \equiv \left(\varphi \metricI{\since} \psi\right) \wedge \left(\chi \metricI{\since} \psi\right)\\
    \metricI{\alwaysF} \left( \varphi \wedge \psi \right) &\equiv \metricI{\alwaysF} \varphi \wedge \metricI{\alwaysF} \psi & \varphi \metricI{\since} \left(\chi \vee \psi\right) &\equiv \left(\varphi \metricI{\since} \chi\right) \vee \left(\varphi \metricI{\since} \psi\right)\\
    \metricI{\wprevious} \left( \varphi \vee \psi \right) &\equiv \metricI{\wprevious} \varphi \vee \metricI{\wprevious} \psi & \metricI{\previous} \left( \varphi \vee \psi \right) &\equiv \metricI{\previous} \varphi \vee \metricI{\previous} \psi\\
    \metricI{\wprevious} \left( \varphi \wedge \psi \right) &\equiv \metricI{\wprevious} \varphi \wedge \metricI{\wprevious} \psi & \metricI{\previous} \left( \varphi \wedge \psi \right) &\equiv \metricI{\previous} \varphi \wedge \metricI{\previous} \psi\\
    \metricI{\eventuallyP} \left( \varphi \vee \psi \right) &\equiv \metricI{\eventuallyP} \varphi \vee \metricI{\eventuallyP} \psi & \metricI{\alwaysP} \left( \varphi \wedge \psi \right) &\equiv \metricI{\alwaysP} \varphi \wedge \metricI{\alwaysP} \psi\\
    \left(\varphi\vee \chi \right) \metricI{\trigger} \psi &\equiv \left(\varphi \metricI{\trigger} \psi\right) \vee \left(\chi \metricI{\trigger} \psi\right) & \varphi \metricI{\trigger} \left(\chi \wedge \psi\right) & \equiv \left(\varphi \metricI{\trigger} \chi\right) \wedge \left(\varphi \metricI{\trigger} \psi\right)
  \end{align*}
\end{proposition}

We can also prove a kind of De~Morgan duality between until and release, and analogously, between since and trigger:
%
\begin{proposition}\label{prop:equiv:negation}
  For all metric formulas $\varphi$ and $\psi$, the following equivalences hold in \MHT:
  \begin{align*}
    \neg \left(\varphi \metricI{\until} \psi \right) &\equiv \neg \varphi \metricI{\release} \neg \psi & \neg \left(\varphi \metricI{\since} \psi \right) &\equiv \neg \varphi \metricI{\trigger} \neg \psi\\
    \neg \left(\varphi \metricI{\release} \psi \right) &\equiv \neg \varphi \metricI{\until} \neg \psi & \neg \left(\varphi \metricI{\trigger} \psi \right) & \equiv \neg \varphi \metricI{\since} \neg \psi
  \end{align*}
\end{proposition}

Another interesting result has to do with the effect of extending or stretching the interval in these operators.
%
\begin{proposition}\label{prop:intervals} Let $\cI$ and $\cJ$ be two intervals satisfying $\cI \subseteq \cJ$.
  For all metric formulas $\varphi$ and $\psi$, the following expressions are valid in \MHT:
  \begin{align*}
    \left(\varphi \until_{\cI} \psi\right) & \rightarrow \left(\varphi \until_{\cJ} \psi\right) & \left(\varphi \release_{\cJ} \psi\right) & \rightarrow \left(\varphi \release_{\cI} \psi\right) \\
    \left(\varphi \since_{\cI} \psi\right) & \rightarrow \left(\varphi \since_{\cJ} \psi\right) & \left(\varphi \trigger_{\cJ} \psi\right) & \rightarrow \left(\varphi \trigger_{\cI} \psi\right)
  \end{align*}
\end{proposition}

We observe next the effect of the semantics of \emph{always} and \emph{eventually} on truth constants.
If $m, n \in \mathbb{N}$,
then
${\alwaysF}_{\intervo{m}{n}} \bot$ means that there is no state in interval $\intervo{m}{n}$
and
${\eventuallyF}_{\intervo{m}{n}} \top$ means that there is at least one state in this interval.
The formula
${\alwaysF}_{\intervo{m}{n}} \top$
is a tautology, whereas
${\eventuallyF}_{\intervo{m}{n}} \bot$
is unsatisfiable.
The same applies to past operators ${\eventuallyP}_{\intervo{m}{n}}$ and ${\alwaysP}_{\intervo{m}{n}}$.

\subsection*{Strict traces}

In the rest of this section, we consider a group of results that hold under the assumption of strict traces, namely,
that $\tmf(i) < \tmf(i+1)$ for any pair of consecutive time points.
We can enforce metric models to be traces with a strict timing function $\tmf$.
This can be achieved with the simple addition of the axiom $\alwaysF \neg \metric{\next}{0}{0} \top$.
In the following, we assume that this axiom is included and consider, in this way, strict timing.
%

The following equivalences state that interval \intervc{0}{0} makes all binary metric operators
collapse into their right hand argument formula,
whereas unary operators collapse to a truth constant.
For metric formulas $\psi$ and $\varphi$ and for strict traces, we have:
\begin{align}
  \psi\metric{\until}{0}{0}\varphi &{} \equiv \psi\metric{\release}{0}{0}   \varphi \equiv \varphi \label{inductive_def_zero:1}
  \\
  \metric{\next}{0}{0} \,  \varphi & {} \equiv \metric{\previous}{0}{0} \,  \varphi \equiv \bot    \label{inductive_def_zero:2}
  \\
  \metric{\wnext}{0}{0} \, \varphi & {} \equiv \metric{\wprevious}{0}{0} \, \varphi \equiv \top    \label{inductive_def_zero:3}
\end{align}
The last two lines are precisely an effect of dealing with strict traces.
For instance, $\metric{\next}{0}{0} \, \varphi \equiv \bot$ tells us that
it is always impossible to have a successor state with the same time (the time difference is 0) as the current one,
regardless of the formula $\varphi$ at hand.

The next lemma allows us to unfold metric operators for single-point time intervals $\intervc{n}{n}$ with $n>0$.
Moreover, this lemma shows that for a single-point time interval $\metric{\next}{n}{n}\varphi$ is definable.
%
\begin{lemma}\label{lemma_inductive_def_point}
  For metric formulas $\psi$ and $\varphi$ and for $n>0$, we have:

  \begin{minipage}[b]{0.45\linewidth}
    \begin{align}\label{def:inductive_next_n}
    	\metric{\next}{n}{n} \varphi & {}\textstyle\equiv {\alwaysF}_{\intervo{1}{n}} \bot \wedge \metric{\eventuallyF}{n}{n}\varphi\\
    	\label{def:inductive_until_n}
      \psi\metric{\until}{n}{n} \varphi
      & {} \equiv\textstyle
      \psi\wedge \bigvee_{i=1}^n \metric{\next}{i}{i} ( \psi\metric{\until}{n{-}i}{n{-}i} \varphi )
      \\\label{def:inductive_release_n}
      \psi\metric{\release}{n}{n} \varphi
      & {} \equiv\textstyle
      \psi \vee \bigwedge_{i=1}^n \metric{\wnext}{i}{i} ( \psi\metric{\release}{n{-}i}{n{-}i} \varphi )
    \end{align}
  \end{minipage}
  \begin{minipage}[b]{0.45\linewidth}
    \begin{align}\label{def:inductive_wnext_n}
    	\wnext_{n} \varphi & {}\textstyle\equiv \eventuallyF_{[1..n)}\top \vee \metric{\alwaysF}{n}{n} \varphi\\
    	\label{def:inductive_eventually_n}
      \metric{\eventuallyF}{n}{n} \varphi
      & {} \equiv\textstyle
      \bigvee_{i=1}^n \metric{\next}{i}{i} \metric{\eventuallyF}{n{-}i}{n{-}i} \varphi
      \\
      \label{def:inductive_always_n}\metric{\alwaysF}{n}{n} \varphi
      & {} \equiv\textstyle
      \bigwedge_{i=1}^n \metric{\wnext}{i}{i} \metric{\alwaysF}{n{-}i}{n{-}i} \varphi
    \end{align}
  \end{minipage}
\qed
\end{lemma}
%
The same applies for the dual past operators.

Going one step further,
we can also unfold \emph{until} and \emph{release} for intervals of the form $\intervc{0}{n}$ with the application of the following result.
%
\begin{lemma}\label{lemma_inductive_def_interval_zero}
  For metric formulas $\psi$ and $\varphi$ and for $n > 0$, we have:
  \begin{align}\label{def:inductive_until_0n}
    \psi \metric{\until}{0}{n} \varphi
    & {} \equiv\textstyle
    \varphi \vee ( \psi \wedge
    \bigvee_{i=1}^{n} \metric{\next}{i}{i} ( \psi \metric{\until}{0}{(n-i)} \varphi ) )
    \\
    \label{def:inductive_release_0n}\psi \metric{\release}{0}{n} \varphi
    & {} \equiv\textstyle
    \varphi \wedge ( \psi \vee
    \bigwedge_{i=1}^{n} \metric{\wnext}{i}{i} ( \psi \metric{\release}{0}{(n-i)} \varphi ) )
  \end{align}
  \qed
\end{lemma}
%
The same applies for the dual past operators.

We can also prove that for any interval of the form $\intervco{m}{n}$, where $n\not=\omega$, the metric versions of \textit{next} and \textit{weak next} (\textit{resp. previous} and \textit{weak previous}) can be defined in terms of the other metric connectives.

\begin{proposition} \label{prop:definability:next} For all $m , n \in \mathbb{N}$, the following equivalences hold
 \begin{align}\label{def:inductive_next}
	\next_{\intervo{m}{n}} \varphi & {} \equiv \textstyle\bigvee_{i=m}^{n{-}1} \metric{\next}{i}{i} \varphi\\
	\label{def:inductive_wnext}
	\wnext_{\intervo{m}{n}} \varphi & {} \equiv \textstyle\bigwedge_{i=m}^{n{-}1} \metric{\wnext}{i}{i} \varphi
\end{align}
\end{proposition}

The same applies for the dual past operators.

Finally, the next theorem contains a pair of equivalences that, when dealing with finite intervals, can be used to recursively unfold \emph{until} and \emph{release} into combinations of \emph{next} with Boolean operators
(an analogous result applies for \emph{since}, \emph{trigger} and \emph{previous} due to temporal duality).
%
\begin{theorem}[Next-unfolding]\label{lemma_inductive_def_interval}%
  For metric formulas $\psi$ and $\varphi$ and for $m, n \in \mathbb{N}$ such that $0 < m < n-1$, we have:
  \begin{align}
  	\label{def:inductive_until}
    \psi \until_{\intervo{m}{n}} \varphi
    & {} \equiv\textstyle
    \psi \wedge \left(\bigvee_{i=1}^{m} \metric{\next}{i}{i} ( \psi \until_{\intervo{m-i}{n-i}} \varphi )
    \vee
    \bigvee_{i=m{+}1}^{n-1} \metric{\next}{i}{i} ( \psi \metric{\until}{0}{(n-1-i)} \varphi )\right)
    \\\label{def:inductive_release}
    \psi \release_{\intervo{m}{n}} \varphi
    & {} \equiv\textstyle
    \psi \vee \left( \bigwedge_{i=1}^{m} \metric{\wnext}{i}{i} ( \psi \release_{\intervo{(m-i)}{(n-i)}} \varphi )
    \wedge
    \bigwedge_{i=m{+}1}^{n-1} \metric{\wnext}{i}{i} ( \psi \metric{\release}{0}{(n-1-i)} \varphi )\right)
  \end{align}
  \qed
\end{theorem}
%
The same applies for the dual past operators.

As an example,
consider the metric formula $\myatom \until_{\intervo{2}{4}} \myatomb$.
\begin{align*}
  \myatom \until_{\intervo{2}{4}} \myatomb
  & {} \equiv\textstyle
  p \wedge \left(\bigvee_{i=1}^{2} \metric{\next}{i}{i} ( \myatom \until_{\intervo{(2-i)}{(4-i)}} \myatomb )
  \vee
  \bigvee_{i=2{+}1}^{3} \metric{\next}{i}{i} ( \myatom \metric{\until}{0}{(3-i)} \myatomb )\right)
  \\
  & {} \equiv
  p \wedge \left(\metric{\next}{1}{1} ( \myatom \until_{\intervo{1}{3}} \myatomb )
  \vee
  \metric{\next}{2}{2} ( \myatom \metric{\until}{0}{1} \myatomb )
  \vee
  \metric{\next}{3}{3} ( \myatom \metric{\until}{0}{0} \myatomb )\right)
  \\
  & {} \equiv
  p \wedge \left(\metric{\next}{1}{1} ( \myatom \until_{\intervo{1}{3}} \myatomb )
  \vee
  \metric{\next}{2}{2} (
    \myatomb \vee ( \myatom \wedge
    \metric{\next}{1}{1} \myatomb )  )
  \vee
  \metric{\next}{3}{3} \myatomb \right) \\
  & {} \equiv
  p \wedge \left(\metric{\next}{1}{1} (
    \metric{\next}{1}{1} ( \myatomb \vee ( \myatom \wedge \metric{\next}{1}{1} \myatomb ) )
  \vee
  \metric{\next}{2}{2} \myatomb
   )
  \vee
  \metric{\next}{2}{2} (
    \myatomb \vee ( \myatom \wedge
    \metric{\next}{1}{1} \myatomb )  )
  \vee
  \metric{\next}{3}{3} \myatomb\right)
\end{align*}

Another useful result that can be applied to unfold metric operators is the following range splitting theorem.
%
\begin{theorem}[Range splitting]\label{corr:past_future_split}
  For metric formulas $\psi$ and $\varphi$,
  \begin{align}
    \label{eq:range:until}\psi \until_{\intervo{m}{n}} \varphi
      & {} \equiv
      (\psi \until_{\intervo{m}{i}} \varphi) \vee (\psi \until_{\intervo{i}{n}} \varphi)
      \quad \quad \text{ for all } i \in \intervo{m}{n}
    \\
    \label{eq:range:release}\psi \release_{\intervo{m}{n}} \varphi
      & {} \equiv
      (\psi \release_{\intervo{m}{i}} \varphi) \wedge (\psi \release_{\intervo{i}{n}} \varphi)
      \quad \quad \text{ for all } i \in \intervo{m}{n}
  \end{align}
The same applies for the dual past operators.\qed
\end{theorem}

A metric formula $\varphi$ is said to be in \emph{unary normal form},
if intervals only affect unary temporal operators,
while binary operators $\until$, $\release$, $\since$, $\trigger$ are only used in their temporal form,
without any attached intervals.
The following proposition, inspired by~\cite{DSouzaT04},
allows us to translate any arbitrary metric formula into unary normal form.
%
\begin{proposition}\label{prop:defin:until}
  For metric formulas $\varphi$ and $\psi$ and for $m$ and $n$ such that $m > 0$,
  the following equivalences hold in \MHT:
  \begin{align*}
    \varphi \until_{\intervco{m}{n}} \psi &\equiv \eventuallyF_{\intervco{m}{n}} \psi \wedge \alwaysF_{\intervco{0}{m}} \left( \varphi \until \left(\varphi \wedge \next \psi\right)\right) & \varphi \until_{\intervco{0}{n}} \psi &\equiv \eventuallyF_{\intervco{0}{n}}\psi \wedge \varphi \until \psi\\
    \varphi \until_{\intervcc{m}{n}} \psi &\equiv \eventuallyF_{\intervcc{m}{n}} \psi \wedge \alwaysF_{\intervco{0}{m}} \left( \varphi \until \left(\varphi \wedge \next \psi\right)\right) & \varphi \until_{\intervcc{0}{n}} \psi &\equiv  \eventuallyF_{\intervcc{0}{n}}\psi \wedge \varphi \until \psi\\
    \varphi \until_{\intervoo{m}{n}} \psi &\equiv \eventuallyF_{\intervoo{m}{n}} \psi \wedge \alwaysF_{\intervcc{0}{m}} \left( \varphi \until \left(\varphi \wedge \next \psi\right)\right) & \varphi \until_{\intervoo{0}{n}} \psi &\equiv  \eventuallyF_{\intervoo{0}{n}}\psi \wedge \varphi \until \left(\varphi \wedge \next \psi\right)\\
    \varphi \until_{\intervoc{m}{n}} \psi &\equiv \eventuallyF_{\intervoc{m}{n}} \psi \wedge \alwaysF_{\intervcc{0}{m}} \left( \varphi \until \left(\varphi \wedge \next \psi\right)\right) & \varphi \until_{\intervoc{0}{n}} \psi &\equiv  \eventuallyF_{\intervoc{0}{n}}\psi  \wedge \varphi \until \left(\varphi \wedge \next \psi\right)\\
    \varphi \release_{\intervco{m}{n}} \psi &\equiv \alwaysF_{\intervco{m}{n}} \psi \vee \eventuallyF_{\intervco{0}{m}} \left( \varphi \release \left(\varphi \vee \wnext \psi\right)\right) & \varphi \release_{\intervco{0}{n}} \psi &\equiv \alwaysF_{\intervco{0}{n}}\psi \vee \varphi \release \psi \\
    \varphi \release_{\intervcc{m}{n}} \psi &\equiv \alwaysF_{\intervcc{m}{n}} \psi \vee \eventuallyF_{\intervco{0}{m}} \left( \varphi \release \left(\varphi \vee \wnext \psi\right)\right) & \varphi \release_{\intervcc{0}{n}} \psi &\equiv  \alwaysF_{\intervcc{0}{n}}\psi \vee \varphi \release \psi\\
    \varphi \release_{\intervoo{m}{n}} \psi &\equiv \alwaysF_{\intervoo{m}{n}} \psi \vee \eventuallyF_{\intervcc{0}{m}} \left( \varphi \release \left(\varphi \vee \wnext \psi\right)\right) & \varphi \release_{\intervoo{0}{n}} \psi &\equiv  \alwaysF_{\intervoo{0}{n}}\psi \vee \varphi \release \left(\varphi \vee \wnext \psi\right)\\
    \varphi \release_{\intervoc{m}{n}} \psi &\equiv \alwaysF_{\intervoc{m}{n}} \psi \vee \eventuallyF_{\intervcc{0}{m}} \left( \varphi \release \left(\varphi \vee \wnext \psi\right)\right) & \varphi \release_{\intervoc{0}{n}} \psi &\equiv  \alwaysF_{\intervoc{0}{n}}\psi  \vee \varphi \release \left(\varphi \vee \wnext \psi\right)\\
    \varphi \since_{\intervco{m}{n}} \psi &\equiv \eventuallyP_{\intervco{m}{n}} \psi \wedge \alwaysP_{\intervco{0}{m}} \left( \varphi \since \left(\varphi \wedge \previous \psi\right)\right) &\varphi \since_{\intervco{0}{n}} \psi &\equiv \eventuallyP_{\intervco{0}{n}}\psi \wedge \varphi \since \psi \\
    \varphi \since_{\intervcc{m}{n}} \psi &\equiv \eventuallyP_{\intervcc{m}{n}} \psi \wedge \alwaysP_{\intervco{0}{m}} \left( \varphi \since \left(\varphi \wedge \previous \psi\right)\right)&			\varphi \since_{\intervcc{0}{n}} \psi &\equiv  \eventuallyP_{\intervcc{0}{n}}\psi \wedge \varphi \since \psi \\
    \varphi \since_{\intervoo{m}{n}} \psi &\equiv \eventuallyP_{\intervoo{m}{n}} \psi \wedge \alwaysP_{\intervcc{0}{m}} \left( \varphi \since \left(\varphi \wedge \previous \psi\right)\right)&			\varphi \since_{\intervoo{0}{n}} \psi &\equiv  \eventuallyP_{\intervoo{0}{n}}\psi \wedge \varphi \since \left(\varphi \wedge \previous \psi\right) \\
    \varphi \since_{\intervoc{m}{n}} \psi &\equiv \eventuallyP_{\intervoc{m}{n}} \psi \wedge \alwaysP_{\intervcc{0}{m}} \left( \varphi \since \left(\varphi \wedge \previous \psi\right)\right)&			\varphi \since_{\intervoc{0}{n}} \psi &\equiv  \eventuallyP_{\intervoc{0}{n}}\psi  \wedge \varphi \since \left(\varphi \wedge \previous \psi\right) \\
    \varphi \trigger_{\intervco{m}{n}} \psi &\equiv \alwaysP_{\intervco{m}{n}} \psi \vee \eventuallyP_{\intervco{0}{m}} \left( \varphi \trigger \left(\varphi \vee \wprevious \psi\right)\right)&			\varphi \trigger_{\intervco{0}{n}} \psi &\equiv \alwaysP_{\intervco{0}{n}}\psi \vee \varphi \trigger \psi\\
    \varphi \trigger_{\intervcc{m}{n}} \psi &\equiv \alwaysP_{\intervcc{m}{n}} \psi \vee \eventuallyP_{\intervco{0}{m}} \left( \varphi \trigger \left(\varphi \vee \wprevious \psi\right)\right)&			\varphi \trigger_{\intervcc{0}{n}} \psi &\equiv  \alwaysP_{\intervcc{0}{n}}\psi \vee \varphi \trigger \psi\\
    \varphi \trigger_{\intervoo{m}{n}} \psi &\equiv \alwaysP_{\intervoo{m}{n}} \psi \vee \eventuallyP_{\intervcc{0}{m}} \left( \varphi \trigger \left(\varphi \vee \wprevious \psi\right)\right)&			\varphi \trigger_{\intervoo{0}{n}} \psi &\equiv  \alwaysP_{\intervoo{0}{n}}\psi \vee \varphi \trigger \left(\varphi \vee \wprevious \psi\right)\\
    \varphi \trigger_{\intervoc{m}{n}} \psi &\equiv \alwaysP_{\intervoc{m}{n}} \psi \vee \eventuallyP_{\intervcc{0}{m}} \left( \varphi \trigger \left(\varphi \vee \wprevious \psi\right)\right)&			\varphi \trigger_{\intervoc{0}{n}} \psi &\equiv  \alwaysP_{\intervoc{0}{n}}\psi  \vee \varphi \trigger \left(\varphi \vee \wprevious \psi\right)\\
  \end{align*}
\end{proposition}
\begin{corollary}
  Any metric formula can be translated into unary normal form (assuming strict traces).
\end{corollary}
%

\section{Metric Equilibrium Logic}\label{sec:mel}

As in traditional Equilibrium Logic~\cite{pearce96a},
non-monotonicity is achieved in \MEL\ by a selection among the \MHT\ models of a theory.
In what follows, we keep assuming the use of strict traces.

\begin{definition}[Metric Equilibrium/Stable Model]
  Let $\mathfrak{S}$ be some set of timed \HT-traces.
  A total timed \HT-trace $(\tuple{\T, \T}, \tau) \in\mathfrak{S}$ is a metric equilibrium model of $\mathfrak{S}$
  iff there is no other \H\ different from \T\ such that $(\tuple{\H, \T}, \tau) \in \mathfrak{S}$.\qed
\end{definition}

%
We talk about metric equilibrium (or metric stable) models of a theory $\Gamma$ when $\mathfrak{S} = \MHT(\Gamma)$, and we write
$\MEL(\Gamma, \lambda)$ and $\MEL(\Gamma)$ to stand for the metric equilibrium models of $\MHT(\Gamma, \lambda)$ and $\MHT(\Gamma)$, respectively.
\emph{Metric Equilibrium Logic} (MEL) is the non-monotonic logic induced by the metric equilibrium models of metric theories.
As before, variants \MELf\ and \MELo\ refer to \MEL\ when restricted to traces of finite and infinite length, respectively.
\begin{observation}
The set of metric equilibrium models of $\Gamma$ can be partitioned on the trace lengths, namely, $\bigcup_{\lambda=0}^{\omega} \MEL(\Gamma,\lambda) = \MEL(\Gamma)$.\qed
\end{observation}

Back to our example, suppose we have the theory $\Gamma$ consisting of formulas
\eqref{ex:traffic:light}-\eqref{ex:traffic:light:push}, viz.\
%
\begin{align*}
\alwaysF ( \mathit{red} \wedge \mathit{green} \to \bot) &&\eqref{ex:traffic:light}\\
\alwaysF ( \neg\mathit{green} \to \mathit{red} ) &&\eqref{ex:traffic:light:default}\\
\alwaysF \big( \mathit{push} \to {\eventuallyF}_{\intervo{1}{15}}(\metric{\alwaysF}{0}{30}\;  \mathit{green}) \big) &&\eqref{ex:traffic:light:push}
\end{align*}
%
In the example, we abbreviate subsets of the set of atoms $\{\mathit{green},\mathit{push},\mathit{red}\}$ as strings formed by their initials: For instance, $pr$ stands for $\{\mathit{push},\mathit{red}\}$.
For the sake of readability, we represent sequences $(T_0, T_1, T_2)$ as $T_0 \cdot T_1 \cdot T_2$.
Consider first the total models of $\Gamma$: the first two rules force one of the two atoms $\mathit{green}$ or $\mathit{red}$ to hold at every state.
Besides, we can choose adding $\mathit{push}$ or not, but if we do so, $\mathit{green}$ should hold later on according to \eqref{ex:traffic:light:push}.
Now, for any total model $(\tuple{\T,\T},\tau),0 \models \Gamma$ where $\mathit{green}$ or $\mathit{push}$ hold at some states,
we can always form $\H$ in which we remove those atoms from all the states and it is not difficult to see that $(\tuple{\H,\T},\tau),0 \models \Gamma$, so $(\tuple{\T,\T},\tau)$ is not in equilibrium.
As a consequence, metric equilibrium models of $\Gamma$ have the form $(\tuple{\T,\T},\tau)$ being
$\T=\tuple{T_i} _{i \in \intervo{0}{\lambda}}$
with $T_i = \{\mathit{red}\}$ for all ${i \in \intervo{0}{\lambda}}$ and any arbitrary strict timing function $\tau$.
To illustrate non-monotonicity, suppose now that we have $\Gamma'= \Gamma \cup \{ \next_{5}\; \mathit{push}\}$ and, for simplicity, consider length $\lambda=3$ and traces of the form $T_0 \cdot T_1 \cdot T_2$.
Again, it is not hard to see that total models with $\mathit{green}$ or $\mathit{push}$ in state $T_0$ are not in equilibrium,
leaving the only option $T_0=\{\mathit{red}\}$.
The same happens for $\mathit{green}$ at $T_1$, so we get $T_1=\{\mathit{push},\mathit{red}\}$ as only candidate for equilibrium model.
However, since $\mathit{push} \in T_1$,
the only possibility to satisfy the consequent of \eqref{ex:traffic:light:push} is having $\mathit{green} \in T_2$.
Again, we can also see that adding $\mathit{push}$ at that state would not be in equilibrium so that the only trace in equilibrium is the total trace formed with $T_0=\{\mathit{red}\}$, $T_1=\{\mathit{push},\mathit{red}\}$ and $T_2=\{\mathit{green}\}$.
As for the timing, $\tau(0)=0$ is fixed, and satisfaction of formula $(\next_{5}\; \mathit{push})$ fixes $\tau(1)=5$.
Then, from \eqref{ex:traffic:light:push} we conclude that $\mathit{green}$ must hold at any moment starting at $t$ between $5+1$ and $5+14$ and is kept true in all states between $t$ and $t+30$ time units, but as $\lambda=2$, this means just $t$.
To sum up, we get 14 metric equilibrium models with $\tau(0)=0$ and $\tau(1)=5$ fixed, but varying $\tau(2)$ between $6$ and $19$.



We close this section by considering strong equivalence.
Two metric theories $\Gamma_1$ and $\Gamma_2$ are \emph{strongly equivalent} when $\MEL(\Gamma_1 \cup \Delta)=\MEL(\Gamma_2 \cup \Delta)$ for any metric theory $\Delta$.
This means that we can safely replace $\Gamma_1$ by $\Gamma_2$ in any common context $\Delta$ and still get the same set of metric equilibrium models.
The following result shows that checking strong equivalence for \MEL\ collapses to regular equivalence in the monotonic logic of \MHT.
\begin{theorem}\label{prop:sequivalence}
Let $\Gamma_1$ and $\Gamma_2$ be two metric temporal theories built over a finite alphabet $\PV$.
Then, $\Gamma_1$ and $\Gamma_2$ are strongly equivalent iff $\Gamma_1$ and $\Gamma_2$ are \MHT-equivalent.
\end{theorem}
%

\section{Translation into Monadic Quantified Here-and-There with Difference Constraints}\label{sec:kamp}

In a similar spirit as the well-known translation of Kamp~\citeyear{kamp68a} from \LTL\ to first-order logic,
we consider a translation from \MHT\ into a first-order version of \HT,
more precisely,
a function-free fragment of the logic of Quantified Here-and-There with static domains (\QHTS\ in~\cite{peaval08a}).
The word \emph{static} means that the first-order domain $D$ is fixed for both worlds, here and there.
We refer to our fragment of \QHTS\ as \emph{monadic \QHT\ with difference constraints} (or \QHTD\ for short).
In this logic, the static domain is a subset $D \subseteq \mathbb{N}$ of the natural numbers containing at least the element $0 \in D$.
Intuitively, $D$ corresponds to the set of relevant time points (i.e.\ those associated with states) considered in each model.
Note that the first state is always associated with time $0 \in D$.

The syntax of \QHTD\ is the same as for first-order logic with several restrictions:
First, there are no functions other than the 0-ary function (or constant) `$0$' always interpreted as the domain element $0$
(when there is no ambiguity, we drop quotes around constant names).
Second, all predicates are monadic except for a family of binary predicates of the form $\peq{\dd}$ with $\dd \in \mathbb{Z}\cup \lbrace\omega \rbrace$ where $\dd$ is understood as part of the predicate name.
For simplicity, we write $x \peq{\dd} y$ instead of ${\peq{\dd}}(x,y)$ and $x \peq{\dd} y \peq{\dd'} z$ to stand for $x \peq{\dd} y \wedge y \peq{\dd'} z$.
Unlike monadic predicates, the interpretation of $x \peq{\dd} y$ is static (it does not vary in worlds here and there)
and intuitively means that the difference $x-y$ in time points is smaller or equal than $\dd$.
A first-order formula $\varphi$ satisfying all these restrictions is called a \emph{first-order metric formula} or \emph{FOM-formula} for short.
A formula is a \emph{sentence} if it contains no free variables.
For instance, we will see that the metric formula \eqref{ex:traffic:light:push} can be equivalently translated into the FOM-sentence:
\begin{align}
\forall x \, (0 \peq{0} x \wedge \mathit{push}(x) \to
\exists y\, (x \peq{-1} y \peq{14} x \wedge \forall z \, (y \peq{0} z \peq{30} y \to \mathit{green}(z) ) ) ) \label{f:FOMpush}
\end{align}
We sometimes handle \emph{partially grounded} FOM sentences where some variables in predicate arguments have been directly replaced by elements from $D$.
For instance, if we represent \eqref{f:FOMpush} as $\forall x \ \varphi(x)$, the expression $\varphi(4)$ stands for:
\begin{align*}
0 \peq{0} 4 \wedge \mathit{push}(4) \to
\exists y\, (4 \peq{-1} y \peq{14} 4 \wedge \forall z \, (y \peq{0} z \peq{30} y \to \mathit{green}(z) ) ) \end{align*}
and corresponds to a partially grounded FOM-sentence where the domain element $4$ is used as predicate argument in atoms $0 \peq{0} 4$ and $\mathit{push}(4)$.

A $\QHTD$-\emph{signature} is simply a set of monadic predicates $\cP$.
Given $D$ as above,
$\Atoms{D,\cP}$ denotes the set of all ground atoms $p(n)$ for every monadic predicate $p \in \cP$ and every $n \in D$.
A $\QHTD$-interpretation for signature $\cP$ has the form $\tuple{D,H,T}$ where $D \subseteq \mathbb{N}$, $0 \in D$ and $H \subseteq T \subseteq \Atoms{D,\cP}$.
%
\begin{definition}[\QHTD-satisfaction;~\cite{peaval08a}]
  A \QHTD-interpretation $\mathcal{M}=\tuple{D,H,T}$ satisfies a (partially grounded) FOM-sentence $\varphi$,
  written $\mathcal{M}\models \varphi$, if the following conditions hold:
  \begin{enumerate}
  \item $\mathcal{M} \models \top$ and $\mathcal{M} \not \models \bot$
  \item $\mathcal{M} \models p(t)$
    iff
    $p(t) \in H$
  \item $\mathcal{M} \models t_1 \peq{\dd} t_2$
    iff
    $t_1-t_2 \le \dd$ with $t_1, t_2 \in D$
  \item $\mathcal{M} \models \varphi \wedge \psi$
    iff
    $\mathcal{M} \models\varphi$ and $\mathcal{M}  \models\psi$
  \item $\mathcal{M} \models \varphi \vee \psi$
    iff
    $\mathcal{M} \models\varphi$ or $\mathcal{M}  \models\psi$
  \item $\mathcal{M} \models \varphi \rightarrow \psi$
    iff
    $\tuple{D, X, T} \not \models \varphi$ or $\tuple{D, X, T}\models\psi$ for $X \in \{H,T\}$
  \item $\mathcal{M} \models \forall\, x\ \varphi(x)$
    iff
    $\mathcal{M} \models \varphi(t)$ for all $t \in D$
  \item $\mathcal{M} \models \exists\, x\ \varphi(x)$
    iff
    $\mathcal{M} \models \varphi(t)$  for some $t \in D$ \qed
  \end{enumerate}
\end{definition}
%
We can read the expression $x \peq{\dd} y$ as just another way of writing the difference constraint $x-y \leq \delta$.
When $\delta$ is an integer,
we may see it as a lower bound $x-\delta \leq y$ for $y$ or as an upper bound $x \leq y+\delta$ for $x$.
For $\delta=\omega$, $x \peq{\omega} y$ is equivalent to $\top$ since it amounts to the comparison $x-y \leq \omega$.
An important observation is that this difference predicate $\peq{\dd}$ satisfies the excluded middle axiom, that is, the following formula is a \QHTD-tautology:
\[
\forall x \, \forall y\, (\ x \peq{\dd} y \vee \neg (x \peq{\dd} y) \ )
\]
for every $\dd \in \mathbb{Z} \cup \{\omega\}$.
We provide next several useful abbreviations:
\[
\begin{array}{rcl@{\hspace{30pt}}rcl}
x \pt{\dd} y & \eqdef & \neg (y \peq{-\dd} x) \\[2pt]
x \leq y & \eqdef & x \peq{0} y  &
x \neq y & \eqdef & \neg (x = y) \\[2pt]
x = y &\eqdef & \left(x \leq y\right) \wedge \left(y \leq x\right) &
x < y & \eqdef & (x \le y)\wedge (x \not = y) \\[2pt]
\end{array}
\]
For any pair $\odot$, $\oplus$ of comparison symbols,
we extend the abbreviation $x \odot y \oplus z$ to stand for the conjunction $x \odot y \wedge y \oplus z$.
Note that the above derived order relation $x \le y$ captures the one used in Kamp's original translation~\cite{kamp68a} for \LTL.

Equilibrium models for first-order theories are defined as in~\cite{peaval08a}.
\begin{definition}[Quantified Equilibrium Model;~\cite{peaval08a}]
  Let $\varphi$ be a first-order formula.
  A total $\QHTD$-interpretation $\tuple{D,T,T}$ is a first-order equilibrium model of $\varphi$ if
  $\tuple{D,T,T} \models \varphi$ and there is no $H \subset T$ satisfying $\tuple{D,H,T} \models \varphi$.
  \qed
\end{definition}

Before presenting our translation,
we need to remark that we consider non-empty intervals of the form $\intervo{m}{n}$ with $m < n$.
%
\begin{definition}[First-order encoding] \label{tel:trans:qht}
  Let $\varphi$ be a metric formula over \PV.
  We define the translation $\tr{\varphi}_x$ of $\varphi$ for some time point $x\in\mathbb{N}$ as follows:
  \begin{eqnarray*}
    \tr{\bot}_x & \eqdef & \bot
    \\
    \tr{\myatom}_x & \eqdef & \myatom(x) \ \text{ for any } \myatom \in \PV
    \\
    \tr{\varphi \otimes \psi}_x & \eqdef & \tr{\varphi}_x \otimes \tr{\beta}_x \ \text{ for any connective } \otimes \in
                                           \{\wedge,\vee,\to\}
    \\
    \tr{\metricMN{\next} \psi}_x & \eqdef & \exists y \left(x < y \wedge \left(\neg \exists z\; x < z < y\right)  \wedge x
                                            \peq{-m} y  \pt{n} x \wedge  \tr{\psi}_y\right)
    \\
    \tr{\metricMN{\wnext} \psi}_x & \eqdef & \forall y \left( x < y \wedge \left(\neg \exists z\; x < z < y \right)  \wedge
                                             x \peq{-m} y  \pt{n} x  \rightarrow  \tr{\psi}_y\right)
    \\
    \tr{\varphi \metricMN{\until} \psi}_{x} & \eqdef &  \exists y\; \left(x \le y \wedge x \peq{-m} y  \pt{n} x \wedge
                                                       \tr{\psi}_y  \wedge \forall z \left(x \le z < y  \rightarrow
                                                       \tr{\varphi}_z\right)\right)
    \\
    \tr{\varphi \metricMN{\release} \psi}_{x} & \eqdef &  \forall y\; \left(\left(x \le y \wedge x \peq{-m} y  \pt{n}
                                                         x\right) \rightarrow \left( \tr{\psi}_y  \vee \exists z \left( x
                                                         \le z < y \wedge \tr{\varphi}_z\right)\right)\right)
    \\
    \tr{\metricMN{\previous} \psi}_x & \eqdef & \exists y \left( y < x \wedge \neg \exists z\; \left(y < z < x \right)
                                                \wedge  x \pt{n} y \peq{-m} x \wedge \tr{\psi}_y\right)
    \\
    \tr{\metricMN{\wprevious} \psi}_x & \eqdef & \forall y \left( \left(y < x \wedge \neg \exists z\; \left( y < z <
                                                 x\right) \wedge  x \pt{n} y \peq{-m} x \right)\rightarrow
                                                 \tr{\psi}_y\right)
    \\
    \tr{\varphi \metricMN{\since} \psi}_x & \eqdef & \exists y\; \left(y \le x  \wedge x \pt{n} y \peq{-m} x \wedge
                                                     \tr{\psi}_y \wedge \forall \left( y < z\le x  \rightarrow
                                                     \tr{\varphi}_z\right)\right)
    \\
    \tr{\varphi \metricMN{\trigger} \psi}_x & \eqdef & \forall y\; \left(\left(y \le x \wedge x \pt{n} y \peq{-m} x\right)
                                                       \rightarrow \left(\tr{\psi}_y \vee  \exists z \left( y < z \le x
                                                       \wedge \tr{\varphi}_z\right)\right)\right)
  \end{eqnarray*}\qed
\end{definition}
%
Each quantification introduces a new variable.
For instance, consider the translation of \eqref{ex:traffic:light:push} at point $x=0$.
Let us denote \eqref{ex:traffic:light:push} as $\alwaysF ( \mathit{push} \to \alpha )$ where $\alpha$ is the formula
${\eventuallyF}_{\intervo{1}{15}}(\metric{\alwaysF}{0}{30}\; \mathit{green})$.
Then, if we translate the outermost operator $\alwaysF$, we get:
\begin{eqnarray*}
& & \tr{\alwaysF ( \mathit{push} \to \alpha ) }_0\\
& = & \tr{\bot \ \release_{\intervo{0}{\omega}} ( \mathit{push} \to \alpha )}_0 \\
& = &\forall y\; \left(\left(0 \le y \wedge 0 \peq{-0} y  \pt{\omega} 0\right) \rightarrow \left( \tr{\mathit{push} \to \alpha}_y  \vee \exists z \left( 0 \le z < y \wedge \bot \right)\right)\right)\\
& \equiv & \forall y\; (0 \le y \wedge 0 \le y \wedge \top \rightarrow ( \tr{\mathit{push}}_y \to \tr{\alpha}_y) \vee \bot )\\
& \equiv & \forall y\; (0 \le y \wedge \mathit{push}(y) \to \tr{\alpha}_y)\\
& \equiv & \forall x\; (0 \le x \wedge \mathit{push}(x) \to \tr{\alpha}_x)
\end{eqnarray*}
where we renamed the quantified variable for convenience.
If we proceed further, with $\alpha$ as ${\eventuallyF}_{\intervo{1}{15}}\beta$
and letting $\beta$ be $(\metric{\alwaysF}{0}{30}\; \mathit{green})$, we obtain:
\begin{eqnarray*}
\tr{\alpha}_x & = & \tr{{\eventuallyF}_{\intervo{1}{15}}\beta}_x\\
& = & \tr{\top \; \until_{\intervo{1}{15}} \; \beta}_x\\
& = & \exists y\; \left(x \le y \wedge x \peq{-1} y  \pt{15} x \wedge \tr{\beta}_y  \wedge \forall z \left(x \le z < y  \rightarrow \top\right)\right)\\
& \equiv & \exists y\; \left(x \peq{-1} y  \pt{15} x \wedge \tr{\beta}_y \right)
\equiv \exists y\; \left(x \peq{-1} y  \peq{14} x \wedge \tr{\beta}_y \right)
\end{eqnarray*}
Finally, the translation of $\beta$ at $y$ amounts to:
\begin{eqnarray*}
& & \tr{\metric{\alwaysF}{0}{30}\;  \mathit{green}}_{y}\\
& = & \tr{\bot \release_{\intervo{0}{30}}\;  \mathit{green}}_{y} \\
& = & \forall y'\; \left(\ y \le y' \wedge y \peq{-0} y'  \pt{30} y \rightarrow \mathit{green}(y')  \vee \exists z \left( y \le z < y' \wedge \bot \right)\ \right)\\
& \equiv & \forall y'\; \left(\ y \le y' \wedge y \peq{0} y' \wedge y' \pt{30} y \rightarrow \mathit{green}(y') \ \right)\\
& \equiv & \forall y'\; \left(\ y \peq{0} y' \pt{30} y \rightarrow \mathit{green}(y') \ \right)\\
& \equiv & \forall z\; \left(\ y \peq{0} z \pt{30} y \rightarrow \mathit{green}(z) \ \right)
\end{eqnarray*}
so that, when joining all steps together, we get the formula \eqref{f:FOMpush} given above.

The following model correspondence between \MHTf\ and \QHTD\ interpretations can be established.
Given a timed trace $(\tuple{\H,\T}, \tau)$ of length $\lambda > 0$ for signature $\PV$,
we define the first-order signature $\cP = \{ \myatom/1 \mid \myatom \in \PV \}$ containing unary predicates
and a corresponding \QHTD\ interpretation $\tuple{D,H,T}$ where
$D=\{\tau(i) \mid i \in \intervo{0}{\lambda}\}$,
$H = \lbrace \myatom(\tau(i)) \mid  i \in \intervo{0}{\lambda} \text{ and } \myatom \in H_i \rbrace$ and
$T = \lbrace \myatom(\tau(i)) \mid  i \in \intervo{0}{\lambda} \text{ and } \myatom \in T_i \rbrace$.
Under the assumption of strict semantics,
the following model correspondence can be proved by structural induction.
%
\begin{theorem}\label{thm:kamp}
  For all metric formulas $\varphi$,
  and for all timed traces $(\tuple{\H,\T}, \tau)$ whose
  corresponding \QHTD\ interpretation is denoted by $\tuple{D,H,T}$ and for all $i \in \intervo{0}{\lambda}$.
  \begin{eqnarray}
    (\tuple{\H,\T}, \tau), i \models \varphi & \text{ iff }&  \tuple{D,H,T} \models \tr{\varphi}_{\tau(i)} \label{thm:kamp:ih1}
  \end{eqnarray}\qed	
\end{theorem}
\section{Related Work}\label{sec:related_work}

Seen from far, we have presented an extension of the logic of Here-and-There with qualitative and quantitative temporal constraints.
More closely, our logics \MHT\ and \MEL\ can be seen as metric extensions of the linear-time logics \THT\ and \TEL\
obtained by constraining temporal operators by intervals over natural numbers.
The current approach generalizes the previous metric extension of \TEL\ from~\cite{cadiscsc20a} by
uncoupling the ordinal position $i$ of a state in the trace from its location in the time line $\tau(i)$, which indicates now the elapsed time since the beginning of that trace.
Thus,
while $\eventuallyF_{\intervc{5}{5}}\; \myatom$ meant in~\cite{cadiscsc20a} that $\myatom$ must hold exactly after 5 transitions,
it means here that there must be some future state (after $n>0$ transitions) satisfying $p$ and located 5 time units later.
As a first approach, we have considered time points as natural numbers, $\tau(i) \in \mathbb{N}$.
Our choice of a discrete rather than continuous time domain is primarily motivated by our practical objective to implement the logic programming fragment of \MEL\ on top of existing temporal ASP systems, like \telingo, and thus to avoid undecidability.
%
%
%
%

The need for quantitative time constraints is well recognized and many metric extensions have been proposed.
For instance, actions with durations are considered in~\cite{sobatu04a} in an action language adapting a state-based approach.
Interestingly, quantitative time constraints also gave rise to combining ASP with Constraint Solving~\cite{baboge05a};
this connection is now semantically reinforced by our translation advocating the enrichment of ASP with difference constraints.
Even earlier, metric extensions of Logic Programming were proposed in~\cite{brzoska93a}. 
As well, metric extensions of Datalog are introduced in~\cite{wagrkaka19a} and applied to stream reasoning in~\cite{wakagr19a}.
An ASP-based approach to stream reasoning is elaborated in abundance in \cite{bedaei18a}.
Streams can be seen as infinite traces.
Hence, apart from certain dedicated concepts, like time windows, such approaches bear a close relation to metric reasoning.
Detailing this relationship is an interesting topic of future research.
More remotely, metric constructs were used in trace alignment~\cite{gimupape21a}, scheduling~\cite{luvalibemc16a}, and an extension to Golog~\cite{hoflak18a}.

%

\section{Conclusion}\label{sec:discussion}

We have developed a metric extension of linear-time temporal equilibrium logic, in which temporal operators are constrained by intervals over natural numbers.
The resulting Metric Equilibrium Logic provides the foundation of an ASP-based approach for specifying qualitative and quantitative dynamic constraints.
This expressiveness is useful whenever planning and scheduling go hand in hand, like for instance when actions have durations and their effects need to meet deadlines.
To the best of our knowledge,
our approach is the first that considers the concept of timed traces in the context of temporal ASP.

As future work,
we aim to implement our approach by exploiting the provided translation of metric formulas into monadic first-order formulas.
We expect that such a translation is tailored for an implementation in terms of ASP modulo difference constraints.
A further line of research will be the investigation of the combination of metric concepts with more expressive logics than $\TEL$.
As a first approach in this direction, we recently proposed an extension of dynamic equilibrium logic to incorporate quantitative temporal constraints, leading to metric dynamic equilibrium logic \cite{becadiscsc23}.
%

\paragraph{Acknowledgments}
This work was supported by
MICINN, Spain, grant PID2020-116201GB-I00, Xunta de Galicia, Spain (GPC ED431B 2019/03),
R{\'e}gion Pays de la Loire, France ({\'e}toiles montantes CTASP),
DFG grants SCHA 550/15, Germany,
and
European Union COST action CA-17124.


\paragraph{Competing interests}
The authors declare none.
\nocite{cadiscsc22a}
\bibliography{krr,procs,local}
\bibliographystyle{acmtrans}
\newpage
\appendix
\section{Proofs}\label{sec:appendix}


\begin{proofof}{Proposition~\ref{prop:satisfaction:tel}}
	\begin{align*}
		&\text{\phantom{ iff }}
		\M,k \models \initially \\
		&\text{ iff } \M,k \models \neg \metric{\previous}{0}{\omega} \top &
		\text{by Definition of } \initially\\
		&\text{ iff } \M,k \models \neg {\previous}_{\intervo{0}{\omega}} \top &
		\text{by Definition of } \previous\\
		&\text{ iff } \M,k \not \models \previous_{\intervo{0}{\omega}} \top &
		\text{by Proposition }\ref{prop:negation} \text{ and }\ref{prop:persistence} \\
		& \text{ iff }
	   \M,k-1 \not \models \top \text{ or } k = 0 \text{ or } \tmf(k)-\tmf(k-1) \not \in \intervo{0}{\omega} &
	   \text{by Definition}~\ref{def:mht:satisfaction}\eqref{def:mhtsat:previous}\\
	   & \text{ iff } k = 0 \text{ or } \tmf(k)-\tmf(k-1) \not \in \intervo{0}{\omega} &
	   \M,k \models \top \text{ for all } \rangeco{k}{0}{\lambda}\\
	   & \text{ iff } k = 0
	   &\text{since } \tmf(k-1) \leq \tmf(k) \\
   \end{align*}

   \begin{align*}
		   &\text{\phantom{ iff }}
		   \M,k \models  \metricI{\wprevious} \varphi \\
		   &\text{ iff } \M,k \models \metricI{\previous} \top \to \metricI{\previous}\varphi
		   &\text{ by Definition of } \metricI{\wprevious}\\
		   &\text{ iff } \M,k \not \models \metricI{\previous} \top \hbox{ or } \M,k \models \metricI{\previous} \varphi &\text{ since }  \metricI{\previous} \top  \hbox{ behaves classically} \\
		   &\text{ iff } k=0 \text{ or }  \tmf(k)-\tmf(k-1) \not \in \cI \text{ or }  \M,k \models \metricI{\previous} \varphi
		   &\text{by the satisfaction of }\metricI{\previous}\top \\
		   &\text{ iff } k=0 \text{ or }  \tmf(k)-\tmf(k-1) \not \in \cI \text{ or } \M,k-1 \models \varphi
		   &\text{by the satisfaction of }\metricI{\previous}\varphi\\
		   & & \hbox{ and some propositional reasoning}.
	   \end{align*}


   \begin{align*}
	   &\text{\phantom{ iff }}
	   \M,k \models \metricI{\eventuallyP} \varphi \\
	   &\text{ iff } \M,k \models \top \metricI{\since} \varphi &
	   \text{by Definition of } \metricI{\eventuallyP}\\
	   & \text{ iff for some } \rangec{i}{0}{k} \text{ with } \tmf(k)-\tmf(i) \in \cI & \\
	   &\quad \text{we have }\M, i \models \varphi \text{ and } \M, j \models \top \text{ for all } \rangeoc{j}{i}{k}
	   &\text{by Definition}~\ref{def:mht:satisfaction}\eqref{def:mhtsat:since}\\
	   &\text{ iff }\M,i \models \varphi \text{ for some } \rangecc{i}{0}{k}
	   \text{ with } \tmf(k)-\tmf(i) \in \cI
	   &\M,k \models \top \text{ for all } \rangeco{k}{0}{\lambda}\\
   \end{align*}

   \begin{align*}
	   &\text{\phantom{ iff }}
	   \M,k \models \metricI{\alwaysP} \varphi \\
	   & \text{ iff } \M,k \models \bot \metricI{\trigger} \varphi
	   &\text{by Definition of } \metricI{\alwaysP}\\
	   & \text{ iff for all } \rangec{i}{0}{k} \text{ with } \tmf(k)-\tmf(i) \in \cI
	   \text{,we have } \M, i \models \varphi \text{ or } &\\
	   &\quad \M,j \models \bot \text{ for some } \orange{j}{i}{k}&
	   \text{by Definition}~\ref{def:mht:satisfaction}\eqref{def:mhtsat:trigger}\\
	   & \text{ iff }
	   \M,i \models \varphi \text{ for all } \rangecc{i}{0}{k}
	   \text{ with } \tmf(k)-\tmf(i) \in \cI
	   &\M,k \not \models \bot \text{ for all } \rangeco{k}{0}{\lambda}\\
   \end{align*}

   For the resp. future cases 16-19 the same reasoning applies.
	\end{proofof}

\begin{proofof}{Proposition~\ref{prop:tht-fragment}}
For the complete definition of \THT\ satisfaction, we refer the reader to~\cite{agcadipescscvi023}.
Here, it suffices to observe that, when we use interval $I=\intervo{0}{\omega}$ in all operators, all conditions $x \in I$ in Definition~\ref{def:mht:satisfaction} (\MHT\ satisfaction) become trivially true, so that the use of $\tau$ is irrelevant and the remaining conditions happen to coincide with \THT\ satisfaction.
\end{proofof}

\begin{proofof}{Proposition~\ref{prop:persistence}}
The proof follows by structural induction on the formula $\varphi$.
Note that universal quantification of $k \in \intervo{0}{\lambda}$ is part of the induction hypothesis.
In what follows, we denote $\M=(\tuple{\H,\T},\tmf)$.
\begin{itemize}
\item If $\varphi=\bot$, the property holds trivially because $\M,k \not\models \bot$.
\item If $\varphi$ is an atom $p$, $\M,k \models p$ implies $p \in H_k \subseteq T_k$ and so $(\tuple{\T,\T},\tmf),k \models p$
\item For conjunction, disjunction and implication the proof follows the same steps as with persistence in (non-temporal) \HT
\item If $\varphi=\metricI{\next} \alpha$ then $k+1<\lambda$, $\tmf(k{+}1)-\tmf(k) \in \cI$ and $\M, k{+}1 \models \alpha$.
By induction, the latter implies $(\tuple{\T,\T},\tmf), k{+}1 \models \alpha$ so we get the conditions to conclude $(\tuple{\T,\T},\tmf),k \models \metricI{\next} \alpha$.
\item If $\varphi=\alpha \metricI{\until} \beta$ then $\M,k \models \alpha \metricI{\until} \beta$ implies that for some $\rangeo{j}{k}{\lambda}$ with $\tmf(j)-\tmf(k) \in \cI$, we have $\M, j \models \beta$ and $\M, i \models \alpha$ for all $\rangeo{i}{k}{j}$.
Since the induction hypothesis applies on any time point, we can apply it to subformulas $\beta$ and $\alpha$ to conclude for some $\rangeo{j}{k}{\lambda}$ with $\tmf(j)-\tmf(k) \in \cI$, we have $(\tuple{\T,\T},\tmf), j \models \beta$ and $(\tuple{\T,\T},\tmf), i \models \alpha$ for all $\rangeo{i}{k}{j}$.
But the latter amounts to $(\tuple{\T,\T},\tmf),k \models \alpha \metricI{\until} \beta$.
\item The proofs for $\metricI{\previous}$ and $\metricI{\since}$ are completely analogous to the two previous steps, respectively.
\end{itemize}
\end{proofof}


\begin{proofof}{Corollary~\ref{cor:decidability}}
	By referring to \MTLf{}-satisfiability as defined in \cite{koymans90a}, it is obvious that \MHTf{}-satisfiability for total traces collapes to \MTLf{}-satisfiability.
	Therefore we claim that a formula $\varphi$ is satisfiable in \MHTf{} iff $\varphi$ is satisfiable in  \MTLf.
	This together with the decidability of \MTLf{}~\cite{ouawor07a} would imply that \MHTf{} is decidable.

	The claim is proved as follows: from left to right, let us assume that $\varphi$ is \MHTf{}-satisfiable.
	Therefore, there exists a \MHTf{} model $(\tuple{\H,\T},\tmf)$ such that $(\tuple{\H,\T},\tmf),0 \models \varphi$.
	By Proposition~\ref{prop:persistence}, $(\tuple{\T,\T}, \tmf),0 \models \varphi$.
	Therefore, $\varphi$ is \MTLf{}-satisfiable.

	Conversely,
	if $\varphi$ is \MTLf{}-satisfiable then there exists a \MTLf{} model $(\T, \tmf)$ such that $(\T,\tmf), 0 \models \varphi$.
	$(\T,\tmf)$
	can be turned into the \MHTf{} model $(\tuple{\T,\T},\tmf)$ satisfying $\varphi$ at $0$.
	Therefore, $\varphi$ is \MHTf{}-satisfiable.
\end{proofof}


\begin{proofof}{Proposition~\ref{prop:negation}}
Note that $(\tuple{\H,\T},\tmf),k \models \neg \varphi$ amounts to $(\tuple{\H,\T},\tmf),k \models \varphi \to \bot$ and the latter is equivalent to $\M, k \not \models \varphi$ or $\M, k \models  \bot$, for both $\M=(\tuple{\H,\T},\tmf)$ and $\M=(\tuple{\T,\T}, \tmf)$.
Since $\M, k \models  \bot$ never holds, we get that this condition is equivalent to both $(\tuple{\H,\T},\tmf), k \not \models \varphi$ and $(\tuple{\T,\T},\tmf), k \not \models \varphi$.
However, by Proposition~\ref{prop:persistence} (persistence), the latter implies the former, so we get that this is just equivalent to $(\tuple{\T,\T},\tmf), k \not \models \varphi$.
\end{proofof}

\begin{proofof}{Proposition~\ref{prop:total}}
From left to right, assume by contradiction that $\H\not=\T$.
By construction of an HT-trace, $H_j\subseteq T_j$ for all $0 \le j < \lambda$, but as $\H \neq \T$, the subset relation must be strict $H_i\subset T_i$ for some
 $0 \le i < \lambda$.
This means that there exists $p \in \PV$ such that $p \in T_i \setminus H_i$.
Therefore, $(\tuple{\H,\T},\tmf) \not \models p \vee \neg p$.
Since $i \ge 0$ and, clearly, $\tau(i)- \tau(0) \in \intervco{0}{\omega}$, we obtain that $(\tuple{\H,\T},\tmf),0 \not \models \metric{\alwaysF}{0}{\omega} \left(p \vee \neg p\right)$.
As a consequence we get  $(\tuple{\H,\T},\tmf),0 \not \models \EM{\PV}$: a contradiction.
Conversely, assume by contradiction that $(\tuple{\H,\T},\tmf), 0 \not \models \EM{\PV}$.
Therefore, there exists $0\le i < \lambda$ such that $\tau(i)-\tau(0) \in \intervco{0}{\omega}$ and $(\tuple{\H,\T},\tmf), i \not \models p \vee \neg p$.
This means that $p \in T_i \setminus H_i$ so $H_i\subset T_i$.
As a consequence, $\H\not = \T$: a contradiction.
\end{proofof}


\begin{proofof}{Proposition~\ref{prop:nonimpl}}
The proof follows similar steps to Proposition~10 in~\cite{agcadipescscvi023} for the non-metric case (and \LTL\ instead of \MTL).
For a proof sketch, note that if no implication or negation is involved, the evaluation of the formula is exclusively performed on trace $\H$, while the there-component $\T$ is never used, becoming irrelevant (we are free to choose any trace $\T \geq \H$).
Thus, checking the equivalence on total traces $(\tuple{\H,\H},\tmf)$ does not lose generality, whereas total traces exactly correspond to \MTL\ satisfaction.
\end{proofof}


\begin{proofof}{Lemma~\ref{TDT}}
The proof follows similar steps to Lemma~2 in~\cite{agcadipescscvi023} for the non-metric case.
Again, we define $\varrho(\M)$ as the timed trace $(\tuple{\H',\T'},\tmf')$ where $H'_i=H_{\lambda-1-i}$ and $T'_i=T_{\lambda-1-i}$ for all $i \in \intervo{0}{\lambda}$.
The only difference here is that we must also ``reverse'' the time function $\tmf$ defining $\tmf'(i)=\tmf(\lambda{-}1)-\tmf(\lambda{-}1{-}i)$ to keep the same relative distances but in reversed order.
Then, the proof follows from the complete temporal symmetry of satisfaction of operators (when the trace is finite).
\end{proofof}

\begin{proofof}{Theorem~\ref{th:temporal_duality}}
The proof follows similar steps to Theorem~3 in~\cite{agcadipescscvi023} for the non-metric case but relying here on Lemma~\ref{TDT} instead.
\end{proofof}


\begin{proofof}{Proposition~\ref{prop:equivalences1}}
	\begin{align*}
		 &\text{\phantom{ iff }}
		 \M,k \models \metricI{\next} \left( \varphi \vee \psi \right)\\
		 & \text{ iff }
		\M,k+1 \models \varphi \vee \psi \text{ and } \tmf(k+1)-\tmf(k) \in \cI &
		\text{by Definition}~\ref{def:mht:satisfaction}\eqref{def:mhtsat:next}\\
		& \text{ iff }
		\left(\M,k+1 \models \varphi \text{ or } \M,k+1 \models \psi \right) \text{ and } \tmf(k+1)-\tmf(k) \in \cI &
		\text{by Definition}~\ref{def:mht:satisfaction}\eqref{def:mhtsat:or}\\
		& \text{ iff }
		\left(\M,k+1 \models \varphi \text{ and } \tmf(k+1)-\tmf(k) \in \cI \right)&
		\text{by Distributivity}\\
		&\quad \text{ or } \left(\M,k+1 \models \psi  \text{ and } \tmf(k+1)-\tmf(k) \in \cI\right) &\\
		& \text{ iff }
		\M,k \models \metricI{\next }\varphi \vee \metricI{\next }\psi &
		\text{by Definition}~\ref{def:mht:satisfaction}\eqref{def:mhtsat:next}\\
	\end{align*}

	\begin{align*}
		&\text{\phantom{ iff }}
		\M,k \models \metricI{\next} \left( \varphi \wedge \psi \right)\\
		& \text{ iff }
		\M,k+1 \models \varphi \wedge \psi \text{ and } \tmf(k+1)-\tmf(k) \in \cI &
		\text{by Definition}~\ref{def:mht:satisfaction}\eqref{def:mhtsat:next}\\
		& \text{ iff }
		\left(\M,k+1 \models \varphi \text{ and } \M,k+1 \models \psi \right) \text{ and } \tmf(k+1)-\tmf(k) \in \cI &
		\text{by Definition}~\ref{def:mht:satisfaction}\eqref{def:mhtsat:and}\\
		& \text{ iff }
		\left(\M,k+1 \models \varphi \text{ and } \tmf(k+1)-\tmf(k) \in \cI \right)&
		\text{by Distributivity}\\
		&\quad \text{ and } \left(\M,k+1 \models \psi  \text{ and } \tmf(k+1)-\tmf(k) \in \cI\right) &\\
		& \text{ iff }
		\M,k \models \metricI{\next }\varphi \wedge \metricI{\next }\psi &
		\text{by Definition}~\ref{def:mht:satisfaction}\eqref{def:mhtsat:next}\\
	\end{align*}

	\begin{align*}
		&\text{\phantom{ iff }}
		\M,k \models \metricI{\wnext} \left( \varphi \vee \psi \right)\\
		& \text{ iff }
		k+1=\lambda \text{ or }\M,k+1 \models \varphi \vee \psi \text{ or } \tmf(k+1)-\tmf(k) \not \in \cI &
		\text{by Proposition}~\ref{prop:satisfaction:tel}\eqref{def:mhtsat:wnext}\\
		& \text{ iff }
		k+1=\lambda \text{ or }
		\left(\M,k+1 \models \varphi \text{ or } \M,k+1 \models \psi \right) \text{ or } \tmf(k+1)-\tmf(k) \not \in \cI &
		\text{by Definition}~\ref{def:mht:satisfaction}\eqref{def:mhtsat:or}\\
		& \text{ iff } \left(
		k+1=\lambda \text{ or }
		\M,k+1 \models \varphi \text{ or } \tmf(k+1)-\tmf(k) \not \in \cI \right)&
		\text{by Distributivity}\\
		& \quad \text{ or } \left( k+1=\lambda \text{ or } \M,k+1 \models \psi \text{ or } \tmf(k+1)-\tmf(k) \not \in \cI\right) &\\
		& \text{ iff }
		\M,k \models \metricI{\wnext }\varphi \vee \metricI{\wnext }\psi &
		\text{by Proposition}~\ref{prop:satisfaction:tel}\eqref{def:mhtsat:wnext}\\
	\end{align*}

	\begin{align*}
		&\text{\phantom{ iff }}
		\M,k \models \metricI{\wnext} \left( \varphi \wedge \psi \right)\\
		& \text{ iff }
		k+1=\lambda \text{ or }\M,k+1 \models \varphi \wedge \psi \text{ or } \tmf(k+1)-\tmf(k) \not \in \cI &
		\text{by Proposition}~\ref{prop:satisfaction:tel}\eqref{def:mhtsat:wnext}\\
		& \text{ iff }
		k+1=\lambda \text{ or }
		\left(\M,k+1 \models \varphi \text{ and } \M,k+1 \models \psi \right) \text{ or } \tmf(k+1)-\tmf(k) \not \in \cI &
		\text{by Definition}~\ref{def:mht:satisfaction}\eqref{def:mhtsat:and}\\
		& \text{ iff } \left(
		k+1=\lambda \text{ or }
		\M,k+1 \models \varphi \text{ or } \tmf(k+1)-\tmf(k) \not \in \cI \right)&
		\text{by Distributivity}\\
		& \quad \text{ and } \left( k+1=\lambda \text{ or } \M,k+1 \models \psi \text{ or } \tmf(k+1)-\tmf(k) \not \in \cI\right) &\\
		& \text{ iff }
		\M,k \models \metricI{\wnext }\varphi \wedge \metricI{\wnext }\psi &
		\text{by Proposition}~\ref{prop:satisfaction:tel}\eqref{def:mhtsat:wnext}\\
	\end{align*}

	\begin{align*}
		&\text{\phantom{ iff }}
		\M,k \models \metricI{\eventuallyF} \left( \varphi \vee \psi \right) \\
		& \text{ iff }
		\M,i \models \varphi \vee \psi \text{ for some } \rangeo{i}{k}{\lambda}
		\text{ with } \tmf(i)-\tmf(k) \in \cI &
		\text{by Definition}~\ref{def:mht:satisfaction}\eqref{def:mhtsat:eventuallyF}\\
		& \text{ iff }
		\left(\M,i \models \varphi \text{ or } \M,i \models \psi \right)
		\text{ for some } \rangeo{i}{k}{\lambda}
		\text{ with } \tmf(i)-\tmf(k) \in \cI &
		\text{by Definition}~\ref{def:mht:satisfaction}\eqref{def:mhtsat:or}\\
		& \text{ iff }
		\left(\M,i \models \varphi \text{ for some } \rangeo{i}{k}{\lambda}
		\text{ with } \tmf(i)-\tmf(k) \in \cI \right)& \text{by Distributivity}\\
		& \quad \text{ or } \left( \M,i \models \psi \text{ for some } \rangeo{i}{k}{\lambda}
		\text{ with } \tmf(i)-\tmf(k) \in \cI \right)&\\
		& \text{ iff }
		\M,k \models \metricI{\eventuallyF}\varphi \vee \metricI{\eventuallyF}\psi &
		\text{by Definition}~\ref{def:mht:satisfaction}\eqref{def:mhtsat:eventuallyF}\\
	\end{align*}

	\begin{align*}
		&\text{\phantom{ iff }}
		\M,k \models \metricI{\alwaysF} \left( \varphi \wedge \psi \right)\\
	    & \text{ iff }
		\M,i \models \varphi \wedge \psi
		\text{ for all } \rangeo{i}{k}{\lambda}
		\text{ with } \tmf(i)-\tmf(k) \in \cI &
		\text{by Definition}~\ref{def:mht:satisfaction}\eqref{def:mhtsat:alwaysF}\\
		& \text{ iff }
		\left(\M,i \models \varphi \text{ and } \M,i \models \psi \right)
		\text{ for all } \rangeo{i}{k}{\lambda}
		\text{ with } \tmf(i)-\tmf(k) \in \cI &
		\text{by Definition}~\ref{def:mht:satisfaction}\eqref{def:mhtsat:and}\\
		& \text{ iff }
		\left(\M,i \models \varphi \text{ for some } \rangeo{i}{k}{\lambda}
		\text{ with } \tmf(i)-\tmf(k) \in \cI \right)& \text{by Distributivity}\\
		& \quad \text{ and } \left( \M,i \models \psi \text{ for some } \rangeo{i}{k}{\lambda}
		\text{ with } \tmf(i)-\tmf(k) \in \cI \right)&\\
		& \text{ iff }
		\M,k \models \metricI{\alwaysF}\varphi \wedge \metricI{\alwaysF}\psi &
		\text{by Definition}~\ref{def:mht:satisfaction}\eqref{def:mhtsat:alwaysF}\\
	\end{align*}

	\begin{align*}
		&\text{\phantom{ iff }}
		\M,k \models \varphi \metricI{\until} \left(\chi \vee \psi\right) \\
		& \text{ iff }
		\M,i \models \chi \vee \psi \text{ for some } \rangeo{i}{k}{\lambda}
		\text{ with } \tmf(i)-\tmf(k) \in \cI \\
		&\quad \text{ and } \M,j \models \varphi \text{ for all } \rangeo{j}{k}{i} &
		\text{by Definition}~\ref{def:mht:satisfaction}\eqref{def:mhtsat:until}\\
		& \text{ iff }
		\left(\M,i \models \chi \text{ or } \M,i \models \psi \right)
		\text{ for some } \rangeo{i}{k}{\lambda}
		\text{ with } \tmf(i)-\tmf(k) \in \cI \\
		&\quad \text{ and } \M,j \models \varphi \text{ for all } \rangeo{j}{k}{i} &
		\text{by Definition}~\ref{def:mht:satisfaction}\eqref{def:mhtsat:and}\\
		& \text{ iff }
		\M,i \models \chi \text{ for some } \rangeo{i}{k}{\lambda}
		\text{ with } \tmf(i)-\tmf(k) \in \cI \text{ or }&
		\text{by Distributivity}\\
		&\quad \M,i \models \psi
		\text{ for some } \rangeo{i}{k}{\lambda}
		\text{ with } \tmf(i)-\tmf(k) \in \cI \\
		&\quad \text{ and } \M,j \models \varphi \text{ for all } \rangeo{j}{k}{i} & \\
		& \text{ iff }
		\M,k \models \left(\varphi \metricI{\until} \chi\right)
		\vee \left(\varphi \metricI{\until} \psi\right)  &
		\text{by Definition}~\ref{def:mht:satisfaction}\eqref{def:mhtsat:until}\\
	\end{align*}

	\begin{align*}
		&\text{\phantom{ iff }}
		\M,k \models \left(\varphi \wedge \chi \right) \metricI{\until} \psi\\
		& \text{ iff }
		\M,i \models \psi \text{ for some } \rangeo{i}{k}{\lambda}
		\text{ with } \tmf(i)-\tmf(k) \in \cI \\
		&\quad \text{ and } \M,j \models \varphi \wedge \psi \text{ for all } \rangeo{j}{k}{i} &
		\text{by Definition}~\ref{def:mht:satisfaction}\eqref{def:mhtsat:until}\\
		& \text{ iff }
		\M,i \models \psi \text{ for some } \rangeo{i}{k}{\lambda}
		\text{ with } \tmf(i)-\tmf(k) \in \cI &
		\text{by Definition}~\ref{def:mht:satisfaction}\eqref{def:mhtsat:and}\\
		&\quad \text{ and } \M,j \models \varphi \text{ and } \M,j \models \psi \text{ for all } \rangeo{j}{k}{i} &\\
		& \text{ iff }
		\M,i \models \psi \text{ for some } \rangeo{i}{k}{\lambda}
		\text{ with } \tmf(i)-\tmf(k) \in \cI & \text{ by Distributivity}\\
		&\quad \text{ and } \M,j \models \varphi \text{ for all } \rangeo{j}{k}{i} \text{ and }&\\
		&\quad \M,i \models \psi \text{ for some } \rangeo{i}{k}{\lambda}
		\text{ with } \tmf(i)-\tmf(k) \in \cI & \\
		&\quad \text{ and } \M,j \models \chi \text{ for all } \rangeo{j}{k}{i} &\\
		& \text{ iff } \M,k \models \left(\varphi \metricI{\until} \psi  \right)
		\text{ and } \M,k \models \left(\chi \metricI{\until} \psi  \right) &
		\text{by Definition}~\ref{def:mht:satisfaction}\eqref{def:mhtsat:until}
	\end{align*}

	\begin{align*}
		&\text{\phantom{ iff }}
		\M,k \models \varphi \metricI{\release} \left(\chi \wedge \psi \right)\\
		 & \text{ iff for all }
		i \text{ with } \tmf(i)-\tmf(k) \in \mathcal{I}, \text{ we have }&
		\text{by Definition}~\ref{def:mht:satisfaction}\eqref{def:mhtsat:release}\\
		&\quad \M,i \models \chi \wedge \psi \text{ or }
		\M,j \models \varphi \text{ for some } \rangeo{j}{k}{i} &\\
		& \text{ iff for all } i \text{ with }
		\tmf(i)-\tmf(k) \in \mathcal{I}, \text{ we have }&\\
		& \quad \M,i \models \chi \text{ and } \M,i \models \psi \text{ or }
		\M,j \models \varphi \text{ for some } \rangeo{j}{k}{i} &
		\text{by Definition}~\ref{def:mht:satisfaction}\eqref{def:mhtsat:and}\\
		& \text{ iff for all }
		i \text{ with } \tmf(i)-\tmf(k) \in \mathcal{I}, \text{ we have }&\\
		& \quad \M,i \models \chi \text{ or }
		\M,j \models \varphi \text{ for some } \rangeo{j}{k}{i} &\\
		&\quad \text{ and iff for all }
		i \text{ with } \tmf(i)-\tmf(k) \in \mathcal{I}, \text{ we have }&\\
		& \quad \M,i \models \psi \text{ or }
		\M,j \models \varphi \text{ for some } \rangeo{j}{k}{i} &
		\text{by Distributivity}\\
		& \text{ iff } \M,k \models \left( \varphi \metricI{\release} \chi \right) \text{ and }
		\left( \varphi \metricI{\release} \psi \right)&
		\text{by Definition}~\ref{def:mht:satisfaction}\eqref{def:mhtsat:release}\\
		& \text{ iff } \M,k \models \left( \varphi \metricI{\release} \chi \right) \wedge
		\left( \varphi \metricI{\release} \psi \right)&
		\text{by Definition}~\ref{def:mht:satisfaction}\eqref{def:mhtsat:and}
	\end{align*}

	\begin{align*}
		&\text{\phantom{ iff }}
		\M,k \models \left(\varphi \vee \chi \right) \metricI{\release}  \psi \\
		& \text{ iff for all }
		i \text{ with } \tmf(i)-\tmf(k) \in \mathcal{I}, \text{ we have }&
		\text{by Definition}~\ref{def:mht:satisfaction}\eqref{def:mhtsat:release}\\
		&\quad \M,i \models  \psi \text{ or }
		\M,j \models \varphi \vee \chi \text{ for some } \rangeo{j}{k}{i} &\\
		& \text{ iff for all } i \text{ with }
		\tmf(i)-\tmf(k) \in \mathcal{I}, \text{ we have }&
		\text{by Definition}~\ref{def:mht:satisfaction}\eqref{def:mhtsat:or}\\
		& \quad \M,i \models \psi \text{ or } \M,j \models \varphi \text{ or }
		\M,j \models \chi \text{ for some } \rangeo{j}{k}{i} &\\
		& \text{ iff for all } i \text{ with }
		\tmf(i)-\tmf(k) \in \mathcal{I}, \text{ we have }&
		\text{by Distributivity}\\
		& \quad \M,i \models \psi \text{ or }
		\M,j \models \varphi \text{ for some } \rangeo{j}{k}{i} \text{ or } &\\
		& \quad \text{ iff for all } i \text{ with }
		\tmf(i)-\tmf(k) \in \mathcal{I}, \text{ we have }&\\
		& \quad \M,i \models \psi \text{ or }
		\M,j \models \chi \text{ for some } \rangeo{j}{k}{i}\\
		& \text{ iff } \M,k \models \left( \varphi \metricI{\release} \psi \right) \text{ or }
		\left( \chi \metricI{\release} \psi \right)&
		\text{by Definition}~\ref{def:mht:satisfaction}\eqref{def:mhtsat:release}\\
		& \text{ iff } \M,k \models \left( \varphi \metricI{\release} \psi \right) \vee
		\left( \chi \metricI{\release} \psi \right)&
		\text{by Definition}~\ref{def:mht:satisfaction}\eqref{def:mhtsat:or}
	\end{align*}

	For the resp. past cases 11-20 the same reasoning applies.
\end{proofof}

\begin{proofof}{Proposition~\ref{prop:equiv:negation}}
We consider the first equivalence.
From left to right, assume towards a contradiction that $(\tuple{\H,\T},\tmf),i \not \models \neg \varphi \metricI{\release}\neg \psi$.
Therefore, there exists $\rangeco{j}{i}{\lambda}$ such that $\tau(j) - \tau(i) \in I$, $(\tuple{\H,\T},\tmf),j\not \models \neg \psi$ and
for all $\rangeco{k}{i}{j}$, $(\tuple{\H,\T},\tmf),k\not \models \neg \varphi$.
By Proposition~\ref{prop:negation}, $(\tuple{\T,\T}, \tmf),j\models \psi$ and $(\tuple{\T,\T}, \tmf),k \models \varphi$ for all $\rangeco{k}{i}{j}$.
By the semantics of the until operator we obtain that $(\tuple{\T,\T}, \tmf),i \models \varphi \metricI{\until} \psi$.
By Proposition~\ref{prop:negation} it follows that $\tuple{\H,\T}, i \not\models \neg \left(\varphi \metricI{\until}\psi\right)$: a contradiction.

From right to left, if $\tuple{\H,\T}, i \not\models \neg \left(\varphi \metricI{\until}\psi\right)$ then, by Proposition~\ref{prop:negation}, $(\tuple{\T,\T}, \tmf),i \models \varphi \metricI{\until}\psi$.
Therefore there exists $\rangeco{j}{i}{\lambda}$ such that $\tau(j) - \tau(i) \in I$, $(\tuple{\T,\T}, \tmf),j \models \psi$ and
for all $\rangeco{k}{i}{j}$, $(\tuple{\T,\T}, \tmf),k \models \varphi$.
Since $(\tuple{\T,\T}, \tmf)$ satisfies the law of excluded middle, it follows that
$(\tuple{\T,\T}, \tmf),j \not \models \neg \psi$ and
for all $\rangeco{k}{i}{j}$, $(\tuple{\T,\T}, \tmf),k \not \models \neg \varphi$.
By the semantics, $(\tuple{\T,\T}, \tmf),i \not \models \neg \varphi \metricI{\release}\neg \psi$.
By persistency, $(\tuple{\H,\T},\tmf)\not \models \neg \varphi \metricI{\release}\neg \psi$.

The remaining equivalences can be verified in a similar way.
\end{proofof}

\begin{proofof}{Proposition~\ref{prop:intervals}}
	\begin{align*}
		\M,k \models \left(\varphi \until_{\cI} \psi\right) & \text{ iff }
		\M,i \models  \psi \text{ for some } \rangeo{i}{k}{\lambda}
		\text{ with } \tmf(i)-\tmf(k) \in \cI \\
		&\quad \text{ and } \M,j \models \varphi \text{ for all } \rangeo{j}{k}{i} &
		\text{by Definition}~\ref{def:mht:satisfaction}\eqref{def:mhtsat:until}\\
		& \text{ implies } 	\M,i \models  \psi \text{ for some } \rangeo{i}{k}{\lambda}
		\text{ with } \tmf(i)-\tmf(k) \in \cJ \\
		&\quad \text{ and } \M,j \models \varphi \text{ for all } \rangeo{j}{k}{i} &
		\text{since }\cI \subseteq \cJ &\\
		& \text{ iff }
		\M,k \models  \left(\varphi \until_{\cJ} \psi\right)
		& \text{by Definition}~\ref{def:mht:satisfaction}\eqref{def:mhtsat:until}
	\end{align*}

	\begin{align*}
		\M,k \models \left(\varphi \release_{\cJ} \psi\right) & \text{ iff for all }
		\rangeo{j}{k}{\lambda} \text{ with } \tmf(i)-\tmf(k) \in \cJ&\\
		& \quad \text{ we have }\M,i \models  \psi \text{ or }
		\M,j \models \varphi \text{ for some } \rangeo{j}{k}{i} &
		\text{by Definition}~\ref{def:mht:satisfaction}\eqref{def:mhtsat:release}\\
		& \text{ implies for all } \rangeo{j}{k}{\lambda}
		\text{ with } \tmf(i)-\tmf(k) \in \cI&\\
		& \quad \text{ we have }\M,i \models  \psi \text{ or }
		\M,j \models \varphi \text{ for some } \rangeo{j}{k}{i} &
		\text{since }\cI \subseteq \cJ \\
		& \text{ iff }
		\M,k \models \left(\varphi \release_{\cI} \psi\right) &
		\text{by Definition}~\ref{def:mht:satisfaction}\eqref{def:mhtsat:release}
	\end{align*}
	The cases 2 and 4 work analogously
\end{proofof}

\begin{proofof}{equivalences~\eqref{inductive_def_zero:1}-\eqref{inductive_def_zero:3}}
\begin{itemize}
	\item Equivalence~\eqref{inductive_def_zero:1}: Take any $i\in [0,\lambda)$.
	 $(\tuple{\H,\T},\tmf), i \models \varphi \metric{\until}{0}{0} \psi$ iff there exists $j \in [i,\lambda)$ such that $\tau(j) - \tau(i) = 0$, $(\tuple{\H,\T},\tmf),j \models \psi$ and for all $i \le k < j$, $(\tuple{\H,\T},\tmf),k \models \varphi$.

	From  $\tau(j) - \tau(i) = 0$ it follows that  $\tau(j) = \tau(i)$.
	Under strict semantics, it follows $j = i$.
	From this we get the iff $(\tuple{\H,\T},\tmf), i \models \psi$.
	Furthermore,

	$(\tuple{\H,\T},\tmf), i \models \varphi \metric{\release}{0}{0} \psi$ iff for all $j \in [i,\lambda)$ if $\tau(j) - \tau(i) = 0$ and $(\tuple{\H,\T},\tmf),j \not \models \psi$ then there exists $i \le k < j$ such that $(\tuple{\H,\T},\tmf),k \models \varphi$.

	From  $\tau(j) - \tau(i) = 0$ it follows that  $\tau(j) = \tau(i)$.
	Under strict semantics, it follows $j = i$.
	From this we get the iff $(\tuple{\H,\T},\tmf), i \models \psi$.

	\item Equivalence~\eqref{inductive_def_zero:2}: For the case of $\metric{\next}{0}{0} \varphi$ we have that $(\tuple{\H,\T},\tmf), i \models \metric{\next}{0}{0}\varphi $ iff $i+1 < \lambda$, $\tau(i+1)-\tau(i) = 0$ and $(\tuple{\H,\T},\tmf), i+1 \models \varphi$.
	Since we are considering strict semantics, we get that $\tau(i+1)-\tau(i) \not=  0$ and we can derive $\bot$.
	We can follow a similar reasoning for the case of $\metric{\previous}{0}{0}\varphi$.

	\item Equivalence~\eqref{inductive_def_zero:3}:  For the case of $\metric{\wnext}{0}{0} \varphi$ we have that $(\tuple{\H,\T},\tmf), i \models \metric{\wnext}{0}{0}\varphi $ iff if $i+1 < \lambda$ and $\tau(i+1)-\tau(i) = 0$ then $(\tuple{\H,\T},\tmf), i+1 \models \varphi$.
	Since we are considering strict semantics, we get that $\tau(i+1)-\tau(i) \not=  0$ and we can derive $\top$.
	We can follow a similar reasoning for the case of $\metric{\wprevious}{0}{0}\varphi$.
\end{itemize}
\end{proofof}

\begin{proofof}{Lemma~\ref{lemma_inductive_def_point}}
	\begin{itemize}
		\item Equivalence~\eqref{def:inductive_next_n}: from left to right, let us assume that $(\tuple{\H,\T},\tmf), i \models \metric{\next}{n}{n}\varphi$.
		Therefore, $(\tuple{\H,\T},\tmf), i+1 \models \varphi$ and $\tmf(i+1) - \tmf(i) = n$.
		From this we conclude that $(\tuple{\H,\T},\tmf), i \models \metric{\eventuallyF}{n}{n}\varphi$. Let us take any $j \ge i$. If $i=j$ then $\tmf(j)-\tmf(i) = 0 \not \in [1..n)$. If $i+1=j$ then $\tmf(j)-\tmf(i) = n \not \in [1..n)$. If $\lambda > j > i+1$ $\tmf(j)-\tmf(i) = \tmf(j)-(\tmf(i+1)-n) = \tmf(j) - \tmf(i+1) + n$.
		Since we are dealing with strict traces, we get that $\tmf(j)-\tmf(i) > n \not \in [1..n)$. Since $j$ was taken arbitrarily, $(\tuple{\H,\T},\tmf), i \models \alwaysF_{[1..n)}\bot$.

		Conversely, let us assume by contradiction that $(\tuple{\H,\T},\tmf), i \models \alwaysF_{[1..n)}\bot \wedge \metric{\eventuallyF}{n}{n} \varphi$ but $(\tuple{\H,\T},\tmf), i \not \models \metric{\next}{n}{n}\varphi$. Therefore, either $\tmf(i+1) - \tmf(i) < n$ or $\tmf(i+1) - \tmf(i) > n$ or $(\tuple{\H,\T},\tmf), i+1\not \models \varphi$. In the first case we get that $(\tuple{\H,\T},\tmf), i\not \models \alwaysF_{[1..n)}\bot$: a contradiction. In the second case we can easily conclude that $(\tuple{\H,\T},\tmf), i \not \models \metric{\eventuallyF}{n}{n}\varphi$: a contradiction. In third case, from $\tmf(i+1) - \tmf(i) = n$ and $(\tuple{\H,\T},\tmf), i+1 \not\models \varphi$ we conclude that $(\tuple{\H,\T},\tmf), i \not\models \metric{\eventuallyF}{n}{n}\varphi$: a contradiction.
		Therefore, $(\tuple{\H,\T},\tmf), i  \models \metric{\next}{n}{n}\varphi$.

		\item Equivalence~\eqref{def:inductive_until_n}: from left to right, if $(\tuple{\H,\T},\tmf), i \models \psi \metric{\until}{n}{n} \varphi$ then there exists $j \in [i,\lambda)$ such that $\tau(j)-\tau(i) = n$, $(\tuple{\H,\T},\tmf),j\models \varphi$ and for all $i \le k < j$, $(\tuple{\H,\T},\tmf), k\models \psi$.
		Since $n > 0$, $\tau(j)-\tau(i) >0$ implies that $j \ge i+1 > i$ and, under strict semantics, $\tau(j) \ge \tau(i+1) > \tau(i)$.
		If we denote by $m \eqdef \tau(i+1) - \tau(i)$, we can conclude that $0 \le m < n$ and, moreover, $\tau(j) - \tau(i+1) = n-m$.
		Therefore, $(\tuple{\H,\T},\tmf), i+1 \models \psi \metric{\until}{n-m}{n-m} \varphi$.
		Since $\tau(i+1) - \tau(i) = m$, it follows $(\tuple{\H,\T},\tmf),i\models \metric{\next}{m}{m} \left(\psi \metric{\until}{n-m}{n-m} \varphi\right)$.
		Since $m$ is bounded we conclude that

			$$(\tuple{\H,\T},\tmf),i\models \bigvee \limits_{0 < m \le n} \metric{\next}{m}{m} \left(\psi \metric{\until}{n-m}{n-m} \varphi\right).$$

		Since $(\tuple{\H,\T},\tmf),i\models \psi$ we get

		$$(\tuple{\H,\T},\tmf),i\models \psi \wedge \bigvee \limits_{0 < m \le n} \metric{\next}{m}{m} \left(\psi \metric{\until}{n-m}{n-m} \varphi\right).$$

		Conversely, if

		$$(\tuple{\H,\T},\tmf),i\models \psi \wedge \bigvee \limits_{0 < m \le n} \metric{\next}{m}{m} \left(\psi \metric{\until}{n-m}{n-m} \varphi\right),$$

		then $(\tuple{\H,\T},\tmf),i\models \psi$ and there exists $0 < m \le n$ such that $(\tuple{\H,\T},\tmf),i\models \metric{\next}{m}{m} \left(\psi \metric{\until}{n-m}{n-m}\varphi\right)$.
		Therefore, $\tau(i+1) - \tau(i) = m$ and $(\tuple{\H,\T},\tmf),i+1\models \psi \metric{\until}{n-m}{n-m}\varphi$.
		Since $(\tuple{\H,\T},\tmf),i+1\models \psi \metric{\until}{n-m}{n-m}\varphi$, there exists $j\ge i+1$ such that $\tau(j) - \tau(i+1) = n-m$, $(\tuple{\H,\T},\tmf),j\models \varphi$ and $(\tuple{\H,\T},\tmf), k \models \psi$ for all $i+1 \le k < j$.
		From $\tau(j) - \tau(i+1) = n-m$ and $\tau(i+1) - \tau(i) = m$ we get that $\tau(j)-\tau(i) = n$.
		Also, since $(\tuple{\H,\T},\tmf),i\models \psi$, it follows that $(\tuple{\H,\T},\tmf), k \models \psi$ for all $i \le k < j$ leading to $(\tuple{\H,\T},\tmf),i \models \psi \metric{\until}{n}{n}\varphi$

		\item Equivalence~\eqref{def:inductive_wnext_n} from right to left, let us assume that $(\tuple{\H,\T},\tmf),i\not\models \metric{\wnext}{n}{n} \varphi$. Therefore $\tmf(i+1)-\tmf(i) = n$ and $(\tuple{\H,\T},\tmf),i+1\not \models\varphi$.
		Since we are considering strict traces, it follows that $(\tuple{\H,\T},\tmf), i\not \models \metric{\alwaysF}{n}{n}\varphi$.
		Let us take $j \ge i$:
		\begin{itemize}
			\item If $j = i$ then $\tmf(j)-\tmf(i) = 0 \not \in [1..n)$.
			\item If $j = i+1$ then $\tmf(j)-\tmf(i) = n \not \in [1..n)$.
			\item If $j > i+1$ then $\tmf(j)-\tmf(i+1) > 0$ because of considering strict traces. Therefore,
			$\tmf(j)-\tmf(i)-n > 0$ so $\tmf(j)-\tmf(i) \not \in [1..n)$ .
		\end{itemize}
		As a consequence, $(\tuple{\H,\T},\tmf), i \not \models \eventuallyF_{[1..n)}\top$: a contradiction.

		From left to right, let us assume towards a contradiction that $(\tuple{\H,\T},\tmf), i\not \models \eventuallyF_{[1..n)}\top$ and $(\tuple{\H,\T},\tmf), i\not \models \metric{\alwaysF}{n}{n}\varphi$. Therefore, it follows that:

		\begin{enumerate}
			\item for all $j\ge i$, $\tmf(j)-\tmf(i) \not \in [1..n)$ and
			\item there exists $k \ge i$ such that $\tmf(k)-\tmf(i) = n$ and $(\tuple{\H,\T},\tmf), k \not \models \varphi$.
		\end{enumerate}
		\noindent We claim that $k = i+1$. Otherwise, if $k = i$ then $\tmf(k)- \tmf(i) = 0 \not = n$ (a contradiction) and, if $k > i+1$,  we have that $\tmf(i+1)-\tmf(i) > 0$ and $\tmf(j) -\tmf(i+1) > 0$ (so $\tmf(i+1) -\tmf(i) < n$): also a contradiction.

		Finally, from $k=i+1$ and $\tmf(i+1) -\tmf(i) = n$ and $(\tuple{\H,\T},\tmf), i+1 \not \models \varphi$ we conclude that $(\tuple{\H,\T},\tmf), i \not \models \metric{\wnext}{n}{n}\varphi$: a contradiction.

		\item Equivalence~\eqref{def:inductive_release_n}: from right to left assume towards a contradiction that $(\tuple{\H,\T},\tmf), i \not \models \psi \metric{\release}{n}{n} \varphi$ then there exists $j \in [i,\lambda)$ such that $\tau(j)-\tau(i) = n$, $(\tuple{\H,\T},\tmf),j\not \models \varphi$ and for all $i \le k < j$, $(\tuple{\H,\T},\tmf), k \not \models \psi$.
		Since $n > 0$, $n=\tau(j)-\tau(i) >0$ implies that $j \ge i+1 > i$ and, under strict semantics, $\tau(j) \ge \tau(i+1) > \tau(i)$.
		If we denote by $m \eqdef \tau(i+1) - \tau(i)$, we can conclude that $0 < m \le n$.
		Furthermore, it follows that $\tau(j) - \tau(i+1) = n-m$.
		Therefore, $(\tuple{\H,\T},\tmf), i+1 \not \models \psi \metric{\release}{n-m}{n-m} \varphi$.
		Since $\tau(i+1) - \tau(i) = m$, it follows $(\tuple{\H,\T},\tmf),i\not\models \metric{\wnext}{m}{m} \left(\psi \metric{\release}{n-m}{n-m} \varphi\right)$.
		Since $m$ is bounded we conclude that

		$$(\tuple{\H,\T},\tmf),i\not \models \bigwedge \limits_{0 < m \le n} \metric{\wnext}{m}{m} \left(\psi \metric{\release}{n-m}{n-m} \varphi\right).$$

		Since $(\tuple{\H,\T},\tmf),i\not \models \psi$ we reach the contradiction

		$$(\tuple{\H,\T},\tmf),i\not\models \psi \vee \bigwedge \limits_{0 < m \le n} \metric{\wnext}{m}{m} \left(\psi \metric{\release}{n-m}{n-m} \varphi\right).$$

		Conversely, assume towards a contradiction that

		$$(\tuple{\H,\T},\tmf),i \not \models \psi \vee \bigwedge \limits_{0 < m \le n} \metric{\wnext}{m}{m} \left(\psi \metric{\release}{n-m}{n-m} \varphi\right),$$

		then $(\tuple{\H,\T},\tmf),i\not\models \psi$ and there exists $0 < m \le n$ such that $(\tuple{\H,\T},\tmf),i\not \models \metric{\wnext}{m}{m} \left(\psi \metric{\release}{n-m}{n-m}\varphi\right)$.
		Therefore, $\tau(i+1) - \tau(i) = m$ and $(\tuple{\H,\T},\tmf),i+1\not \models \psi \metric{\release}{n-m}{n-m}\varphi$.
		Since $(\tuple{\H,\T},\tmf),i+1\not \models \psi \metric{\until}{n-m}{n-m}\varphi$, there exists $j\ge i+1$ such that $\tau(j) - \tau(i+1) = n-m$, $(\tuple{\H,\T},\tmf),j\not \models \varphi$ and $(\tuple{\H,\T},\tmf), k \not \models \psi$ for all $i+1 \le k < j$.
		From $\tau(j) - \tau(i+1) = n-m$ and $\tau(i+1) - \tau(i) = m$ we get that $\tau(j)-\tau(i) = n$.
		Also, since $(\tuple{\H,\T},\tmf),i\not \models \psi$, it follows that $(\tuple{\H,\T},\tmf), k \not \models \psi$ for all $i \le k < j$ leading to $(\tuple{\H,\T},\tmf),i \not \models \psi \metric{\release}{n}{n}\varphi$: a contradiction.

		\item Equivalences~\eqref{def:inductive_eventually_n}-\eqref{def:inductive_always_n}: by definition, $\metric{\eventuallyF}{n}{n} \varphi \eqdef \top \metric{\until}{n}{n}\varphi$ and $\metric{\alwaysF}{n}{n} \varphi \eqdef \bot \metric{\release}{n}{n}\varphi$.
		Therefore the proof follows directly from equivalences~\eqref{def:inductive_until_n} and~\eqref{def:inductive_release_n}.
	\end{itemize}
\end{proofof}

\begin{proofof}{Lemma~\ref{lemma_inductive_def_interval_zero}}
	\begin{itemize}
		\item Equivalence~\eqref{def:inductive_until_0n}: from left to right, assume $(\tuple{\H,\T},\tmf),k\models \psi \metric{\until}{0}{n} \varphi$, then there is $i \geq k$ with $\tau(i)-\tau(k) \leq n$ s.t. $(\tuple{\H,\T},\tmf),i \models \varphi$ and $(\tuple{\H,\T},\tmf),j \models \psi$ for all $k \leq j < i$.
		Lets further assume towards a contradiction that $(\tuple{\H,\T},\tmf),k \not \models
		\varphi \vee ( \psi \wedge \bigvee_{i=1}^{n} \metric{\next}{i}{i} ( \psi \metric{\until}{0}{(n-i)} \varphi ) ) $
		This implies that $(\tuple{\H,\T},\tmf),k \not \models \varphi$ and $(\tuple{\H,\T},\tmf),k \not \models \psi \wedge \bigvee_{i=1}^{n} \metric{\next}{i}{i} ( \psi \metric{\until}{0}{(n-i)} \varphi )$.
		For the latter to hold either $(\tuple{\H,\T},\tmf),k \not \models \psi$ or $(\tuple{\H,\T},\tmf),k \not \models \bigvee_{i=1}^{n} \metric{\next}{i}{i} ( \psi \metric{\until}{0}{(n-i)} \varphi )$.
		Lets consider $(\tuple{\H,\T},\tmf),k \not \models \psi$ first. To be consistent with the original assumption $(\tuple{\H,\T},\tmf),k \models \varphi$ is needed.
		Since $(\tuple{\H,\T},\tmf),k \not \models \varphi$ was already derived, this leads to a contradiction.
		Considering $(\tuple{\H,\T},\tmf),k \not \models \bigvee_{i=1}^{n} \metric{\next}{i}{i} ( \psi \metric{\until}{0}{(n-i)} \varphi )$ leads to $(\tuple{\H,\T},\tmf),k \not \models \bigvee_{i=1}^{n} \metric{\next}{i}{i} ( \psi \metric{\until}{0}{(n-i)} \varphi )$.
		This implies that there is no $ i > k$ with $ \tau(i)-\tau(k) \leq n$ s.t. $(\tuple{\H,\T},\tmf),i \models \varphi$.
		Together with $(\tuple{\H,\T},\tmf),k \not \models \varphi$ this implies that there is no occurence of $\varphi$ in the interval which is contradictory to the original assumption.

		From right to left, assume $(\tuple{\H,\T},\tmf),k \models
		\varphi \vee ( \psi \wedge \bigvee_{i=1}^{n} \metric{\next}{i}{i} ( \psi \metric{\until}{0}{(n-i)} \varphi ) ) $, then
		$(\tuple{\H,\T},\tmf),k \models \varphi$ or $(\tuple{\H,\T},\tmf),k \models \psi \wedge \bigvee_{i=1}^{n} \metric{\next}{i}{i} ( \psi \metric{\until}{0}{(n-i)} \varphi ) $.
		If $(\tuple{\H,\T},\tmf),k \models \varphi$ then obviously $(\tuple{\H,\T},\tmf),k \models \psi \metric{\until}{0}{n} \varphi$.
		From the second disjunct we get that $(\tuple{\H,\T},\tmf),k \models \metric{\next}{i}{i} ( \psi \metric{\until}{0}{(n-i)} \varphi ) $ for some $1 \leq i \leq n$.
		Then there is a next state that satisfies $\psi \until \varphi$ s.t. $\varphi$ holds somewhere within the next interval and, due to $(\tuple{\H,\T},\tmf),k \models \psi$, $\psi$ holds until then. This implies $(\tuple{\H,\T},\tmf),k \models \psi \metric{\until}{0}{n} \varphi$.
	\end{itemize}

	\begin{itemize}
		\item Equivalence~\eqref{def:inductive_release_0n} follows directly from Equivalence~\eqref{def:inductive_until_0n}, Corollary \ref{BDT} (Boolean Duality) and uniform substitution.
	\end{itemize}
\end{proofof}

\begin{proofof}{Proposition~\ref{prop:definability:next}}
	\begin{itemize}
		\item Equivalence~\eqref{def:inductive_next}: In case $m \ge n$ then $[m \cdots n)$ is empty so $\bigvee_{i=m}^{n{-}1} \metric{\next}{i}{i} \varphi \equiv \bot$. It follows that $\M, k \not \models \metric{\next}_{[m\cdots n)}\varphi $ and $\M, k \not \models \bot$. Otherwise, from left to right, if $\M, k \not \models \metric{\next}_{[m\cdots n)}\varphi $ then $\M, k+1 \models \varphi$ and $\tmf(k+1)-\tmf(k) \in [m \cdots n)$. This means that there exists $\rangeco{t}{m}{n}$ such that $\tmf(k+1)-\tmf(k)=t$. From this and $\M, k+1 \models \varphi$  we get that $\M, k \models \metric{\next}{t}{t}\varphi$ so $\M, k \models \bigvee_{i=m}^{n{-}1} \metric{\next}{i}{i} \varphi$.
		Conversely, if $\M, k \models \bigvee_{i=m}^{n{-}1} \metric{\next}{i}{i} \varphi$ then there exists $\rangeco{t}{m}{n}$ such that $\M, k \models \metric{\next}{t}{t}$. Therefore, $\M, k+1 \models \varphi$ and $\tmf(k+1) -\tmf(k) = t$. Since $\rangeco{t}{m}{n}$ and $\tmf(k+1) -\tmf(k) = t$, $\tmf(k+1) -\tmf(k) \in [m \cdots n)$ so $\M, k \models \next_{[m\cdots n)}\varphi$.

		\item Equivalence~\eqref{def:inductive_wnext}: In case $m \ge n$ then $[m \cdots n)$ is empty so $\bigwedge_{i=m}^{n{-}1} \metric{\wnext}{i}{i} \varphi \equiv \top$. It follows that $\M, k  \models \metric{\wnext}_{[m\cdots n)}\varphi $ and $\M, i \models \top$. Otherwise, from left to right, if $\M, k \not \models \bigwedge_{i=m}^{n{-}1} \metric{\wnext}{i}{i} \varphi$ then there exists $\rangeco{t}{m}{n}$ such that $\M, k \not \models \metric{\wnext}{t}{t}$. Therefore, $\M, k+1 \not \models \varphi$ and $\tmf(k+1) -\tmf(k) = t$. Since $\rangeco{t}{m}{n}$ and $\tmf(k+1) -\tmf(k) = t$, $\tmf(k+1) -\tmf(k) \in [m \cdots n)$ so $\M, k \not \models \wnext_{[m\cdots n)}\varphi$: a contradiction. Conversely, if $\M, k \not \models \metric{\wnext}_{[m\cdots n)}\varphi $ then $\M, k+1 \not \models \varphi$ and $\tmf(k+1)-\tmf(k) \in [m \cdots n)$.
		This means that there exists $\rangeco{t}{m}{n}$ such that $\tmf(k+1)-\tmf(k)=t$.
		From this and $\M, k+1 \not \models \varphi$  we get that $\M, k \not \models \metric{\wnext}{t}{t} \varphi$. Therefore, $\M, k \not \models \bigwedge_{i=m}^{n{-}1} \metric{\wnext}{i}{i} \varphi$: a contradiction.
	\end{itemize}
\end{proofof}

\begin{proofof}{Theorem~\ref{lemma_inductive_def_interval}}
	\begin{itemize}
		\item Equivalence~\eqref{def:inductive_until}:
		from left to right, if $\M, k \models \psi \until_{\intervo{m}{n}} \varphi$
		with the restriction $m>0$ and $m< n -1$, then by Definition ~\ref{def:mht:satisfaction}\eqref{def:mhtsat:until}
		$\M,i \models \varphi$ for some $i \in \intervoo{k}{\lambda}$ with $\tmf(i)-\tmf(k) \in \cI$ and
		$\M,j \models \varphi $ for all $\rangeo{j}{k}{i}$. Since $i>k$ and $\psi$ has to hold since $k$, it follows that $\M,k \models \psi$
		and therefore we get: $\M,k \models \psi$ and  $\M,i \models \varphi$ for some $i \in \intervoo{k}{\lambda}$ with $\tmf(i)-\tmf(k) \in \cI$ and
		$\M,j \models \varphi $ for all $j \in \intervoo{k}{i}$. Notice that $j \in \intervoo{k}{i}$ since $k$ was seperated by $\M,k \models \psi$.
		Now, considering the two options for the distance of the next state: $\tau(k+1) - \tau(k) \leq m$ or $\tau(k+1) - \tau(k) \in \intervoo{m}{n}$, we get:
		$\M,k \models \psi$ and $\M,k+1 \models \psi \until_{\intervo{(m-p)}{(n-p)}} \varphi$ if $p \leq m$ or
		$\M,k+1 \models \psi \until_{<(n-p)} \varphi$ if $p \in \intervoo{m}{n}$, where $ p=\tau(k+1) - \tau(k)$.
		The case $p \leq m$ with $\M,k+1 \models \psi \until_{\intervo{(m-p)}{(n-p)}}$ can be expressed by:

		\begin{equation*}
		\bigvee \limits_{1 \leq i \le m} \metric{\next}{m}{m} \left(\psi \metric{\until}{m-i}{n-i} \varphi\right)
		\end{equation*}

		Similarly the case $p=\tau(k+1) - \tau(k)$ can be represented by $$\bigvee \limits_{m+1 \leq i \le n-1} \metric{\next}{i}{i} \left(\psi \until_{\leq (n-1-i)} \varphi\right)$$
		Taking into account those two options and the already performed conclusion\\ $\M,k \models \psi$, we arrive at:

		\begin{equation*}
		 \M,k \models \psi \wedge \bigvee \limits_{1 \leq i \le m} \metric{\next}{i}{i} \left(\psi \metric{\until}{m-i}{n-i} \varphi\right)
		\vee \bigvee \limits_{m+1 \leq i \le n-1} \metric{\next}{i}{i} \left(\psi \until_{\leq (n-1-i)} \varphi\right)
		\end{equation*}

		Conversely, if

		\begin{equation*}
		\M,k \models \psi \wedge \bigvee \limits_{1 \leq i \le m} \metric{\next}{i}{i} \left(\psi \metric{\until}{m-i}{n-i} \varphi\right)
		\vee \bigvee \limits_{m+1 \leq i \le n-1} \metric{\next}{i}{i} \left(\psi \until_{\leq (n-1-i)} \varphi\right),
		\end{equation*}

		then it follows that $\M,k \models \psi$ and $\M,k+1 \models \psi \until_{\intervco{m-i}{n-i}} \varphi$ for some $i \in \intervcc{1}{m}$
		or $\M,k \models \psi$ and $\M,k+1 \models \psi \until_{\leq (n-1-i)}\varphi$ for some $i \in \intervoo{1}{m}$ since each of the both disjunctions would
		be satisfied in case of one matching next-formula with the respective $i$. From this, together with the assumption $m>0$ and $m<n-1$
		we can conclude that $\M,k \models \psi$ and $\M,i \models \varphi$ for some $i \in \intervoo{k}{\lambda}$
		with $\tau(i)-\tau(k) \in \intervco{m}{n}$ and $\M,j \models \psi$ for all $j \in \intervoo{k}{i}$.
		Finally, by applying Definition~\ref{def:mht:satisfaction}\eqref{def:mhtsat:until} we arrive at:
		$\M, i \models \psi \until_{\intervo{m}{n}} \varphi$

		\item Equivalence~\eqref{def:inductive_release}:
		from left to right, if $\M,k \models \psi \release_{\intervo{m}{n}} \varphi$, then by Definition
		~\ref{def:mht:satisfaction}\eqref{def:mhtsat:release} for all $i \in \intervco{k}{\lambda}$ with
		$\tmf(i)-\tmf(k) \in \intervco{m}{n}$ we have $\M,i \models \varphi$ or $\M,j \models \psi$ for some
		$j \in \intervco{k}{i}$. Let's consider the easy case first, where $\M,k \models \psi$. In this case it is obvious to see that
		$\M,k \models \psi \vee \left( \bigwedge_{i=1}^{m} \metric{\wnext}{i}{i} ( \psi \release_{\intervo{(m-i)}{(n-i)}} \varphi )
		\wedge \bigwedge_{i=m{+}1}^{n-1} \metric{\wnext}{i}{i} ( \psi \metric{\release}{0}{(n-1-i)} \varphi )\right)$.
		If $\M,k \not \models \psi$ there has to be a later releasing $\psi$ for all occurrences of $\neg \psi \in \intervco{m}{n}$.
		In this case the next state, if there is one before the end of the interval, satisfies a Release formula that considers the
		time elapsed since $k$. If the linked time point of the next state is $\in \intervoc{k}{m}$, this state satisfies
		$\bigwedge_{i=1}^{m} \metric{\wnext}{i}{i} ( \psi \release_{\intervo{(m-i)}{(n-i)}} \varphi )$. If the linked time point
		of the next state is $\in \intervoo{m}{n}$, the next state would satisfy
		$\bigwedge_{i=m{+}1}^{n-1} \metric{\wnext}{i}{i} ( \psi \metric{\release}{0}{(n-1-i)} \varphi )$. Considering the possibility
		of both cases we can conclude that
		$\M,k \models \psi \vee \left( \bigwedge_{i=1}^{m} \metric{\wnext}{i}{i} ( \psi \release_{\intervo{(m-i)}{(n-i)}} \varphi )
		\wedge \bigwedge_{i=m{+}1}^{n-1} \metric{\wnext}{i}{i} ( \psi \metric{\release}{0}{(n-1-i)} \varphi )\right)$. \\
		Conversely, assume
		$\M,k \models \psi \vee \left( \bigwedge_{i=1}^{m} \metric{\wnext}{i}{i} ( \psi \release_{\intervo{(m-i)}{(n-i)}} \varphi )
		\wedge \bigwedge_{i=m{+}1}^{n-1} \metric{\wnext}{i}{i} ( \psi \metric{\release}{0}{(n-1-i)} \varphi )\right)$. Again, if
		$\M,k \models \psi$ it follows directly that $\M,k \models \psi \release_{\intervo{m}{n}} \varphi$. \\ If
		$\M,k \models \bigwedge_{i=1}^{m} \metric{\wnext}{i}{i} ( \psi \release_{\intervo{(m-i)}{(n-i)}} \varphi )
		\wedge \bigwedge_{i=m{+}1}^{n-1} \metric{\wnext}{i}{i} ( \psi \metric{\release}{0}{(n-1-i)} \varphi )$, we can also follow that
		$\M,k \models \psi \release_{\intervo{m}{n}} \varphi$, since both the satisfaction of
		$\bigwedge_{i=1}^{m} \metric{\wnext}{i}{i} ( \psi \release_{\intervo{(m-i)}{(n-i)}} \varphi )$ in a next state $\in \intervoc{k}{m}$
		and the satisfaction of $\bigwedge_{i=m{+}1}^{n-1} \metric{\wnext}{i}{i} ( \psi \metric{\release}{0}{(n-1-i)} \varphi )$ in a next state
		$\in \intervoo{m}{n}$ would guarantee it. Taking all possible cases together, we finally arrive at
		$\M,k \models \psi \release_{\intervco{m}{n}} \varphi$.
	\end{itemize}
\end{proofof}

\begin{proofof}{Theorem~\ref{corr:past_future_split}}
	\begin{itemize}
		\item Equivalence~\eqref{eq:range:until}: If $\M,k \models \psi \until_{\intervo{m}{n}} \varphi$, it follows by applying Definition ~\ref{def:mht:satisfaction}\eqref{def:mhtsat:until}
		that $\M,i \models \varphi$ for some $i \in \intervco{k}{\lambda}$ with $\tmf(i)-\tmf(k) \in \intervco{m}{n}$ and $\M,j \models \psi $ for all $\rangeo{j}{k}{i}$.
		From $\tmf(i)-\tmf(k) \in \intervco{m}{n}$ we get $\tmf(i)-\tmf(k) \in \intervco{m}{s}$ or $\tmf(i)-\tmf(k) \in \intervco{s}{n}$ for all $s \in \intervco{m}{n}$.
		This implies that $\M,i \models \varphi$ for some $i \in \intervco{k}{\lambda}$ with $\tmf(i)-\tmf(k) \in \intervco{m}{s}$ and $\M,j \models \psi $ for all $\rangeo{j}{k}{i}$
		or $\M,i \models \varphi$ for some $i \in \intervco{k}{\lambda}$ with $\tmf(i)-\tmf(k) \in \intervco{s}{n}$ and $\M,j \models \psi $ for all $\rangeo{j}{k}{i}$,
		for all $s \in \intervco{m}{n}$. Applying again Definition ~\ref{def:mht:satisfaction}\eqref{def:mhtsat:until} and Definition
		~\ref{def:mht:satisfaction}\eqref{def:mhtsat:or} it follows that
		$\M,k \models \left( \psi \until_{\intervo{m}{i}} \varphi \right) \vee \left( \psi \until_{\intervo{i}{n}} \varphi \right) $
		for all $\rangeo{i}{m}{n}$. As every step of the proof works in the converse direction as well, both directions are provided.
		\item For the case of Equivalence~\ref{eq:range:release} the same reasoning applies.
	\end{itemize}
\end{proofof}

\begin{proofof}{Proposition~\ref{prop:defin:until}}
We assume that we are dealing with strict traces.
We consider first the equivalence $\varphi \until_{\intervco{m}{n}} \psi \equiv \eventuallyF_{\intervco{m}{n}} \psi \wedge \alwaysF_{\intervco{0}{m}} \left( \varphi \until \left(\varphi \wedge \next \psi\right)\right)$.

From left to right, if $(\tuple{\H,\T},\tmf), i \models \varphi \until_{\intervco{m}{n}} \psi$ then there exists $j \ge i$ such that $\rangeco{\tau(j)-\tau(i)}{m}{n}$, $(\tuple{\H,\T},\tmf), j \models \psi$ and for all $\rangeco{k}{i}{j}$, $(\tuple{\H,\T},\tmf), k \models \varphi$.
From  $\rangeco{\tau(j)-\tau(i)}{m}{n}$, $j\ge i$ and $(\tuple{\H,\T},\tmf), j \models \psi$ it follows that $(\tuple{\H,\T},\tmf), i \models \eventuallyF_{\intervco{m}{n}} \psi$.
Moreover, since $m \not = 0$, $\tau(j) - \tau(i) \not= 0$ so $j \not = i$, which implies that $j > i$. 
As a consequence $(\tuple{\H,\T},\tmf), j-1 \models \varphi \wedge \next \psi$ and $(\tuple{\H,\T},\tmf), t \models \varphi$ for all $i \le t < j-1$.

Take any arbitrary $y \ge i$. 
If $y \ge j$ then $\tau(y) - \tau(i) \ge m$ because $\tau(y) \ge \tau(j)$ and $\tau(j)-\tau(i) \ge m$. Therefore, $\tau(y) - \tau(i) \not \in \intervco{0}{m}$.
If $y < j$ then $(\tuple{\H,\T},\tmf),y \models \varphi \until \left(\varphi \wedge \next \psi\right)$.
Since $y$ was arbitrary chosen, it follows that $(\tuple{\H,\T},\tmf), i \models \alwaysF_{\intervco{0}{m}}\left(  \varphi \until \left(\varphi \wedge \next \psi\right)\right)$.
 
For the converse direction, from $(\tuple{\H,\T},\tmf), i \models \alwaysF_{\intervco{0}{m}} \left( \varphi \until \left(\varphi \wedge \next \psi\right)\right)$ it follows that there exists $j > i$ such that $\tau(j)-\tau(i) \ge m$, $(\tuple{\H,\T},\tmf), j \models \psi$ and $(\tuple{\H,\T},\tmf), k \models \varphi$ for all $k\in \intervco{i}{j}$.
Since $(\tuple{\H,\T},\tmf), i \models  \eventuallyF_{\intervco{m}{n}} \psi$ there exists $j'> i$ such that $\tau(j') - \tau(i) \ge m$, $\tau(j') - \tau(i) < n$ and $(\tuple{\H,\T},\tmf),j' \models \psi$.

If $j' < j$ we can easily conclude that $(\tuple{\H,\T},\tmf), i \models \varphi \until_{\intervco{m}{n}} \psi$.
If $j' \ge j$ then $\tau(j') \ge \tau(j)$. 
Since $\tau(j') - \tau(i) < n $ and $\tau(j') > \tau(j)$ $\tau(j)-\tau(i) < n$ so $\tau(j) - \tau(i) \in \intervco{m}{n}$, which leads to $(\tuple{\H,\T},\tmf),i \models \varphi \until_{\intervco{m}{n}} \psi$.\\

\noindent For the second equivalence $\varphi \until_{\intervcc{m}{n}} \psi \equiv \eventuallyF_{\intervcc{m}{n}} \psi \wedge \alwaysF_{\intervco{0}{m}} \left( \varphi \until \left(\varphi \wedge \next \psi\right)\right)$,\\
from left to right, if $(\tuple{\H,\T},\tmf), i \models \varphi \until_{\intervcc{m}{n}} \psi$ then there exists $j \ge i$ such that $\rangecc{\tau(j)-\tau(i)}{m}{n}$, $(\tuple{\H,\T},\tmf), j \models \psi$ and for all $\rangeco{k}{i}{j}$, $(\tuple{\H,\T},\tmf), k \models \varphi$.
From  $\rangecc{\tau(j)-\tau(i)}{m}{n}$, $j\ge i$ and $(\tuple{\H,\T},\tmf), j \models \psi$ it follows that $(\tuple{\H,\T},\tmf), i \models \eventuallyF_{\intervcc{m}{n}} \psi$.
Moreover, since $m \not = 0$, $\tau(j) - \tau(i) \not= 0$ so $j \not = i$, which implies that $j > i$. 
As a consequence $(\tuple{\H,\T},\tmf), j-1 \models \varphi \wedge \next \psi$ and $(\tuple{\H,\T},\tmf), t \models \varphi$ for all $i \le t < j-1$.

Take any arbitrary $y \ge i$. 
If $y \ge j$ then $\tau(y) - \tau(i) \ge m$ because $\tau(y) \ge \tau(j)$ and $\tau(j)-\tau(i) \ge m$. Therefore, $\tau(y) - \tau(i) \not \in \intervco{0}{m}$.
If $y < j$ then $(\tuple{\H,\T},\tmf),y \models \varphi \until \left(\varphi \wedge \next \psi\right)$.
Since $y$ was arbitrary chosen, it follows that $(\tuple{\H,\T},\tmf), i \models \alwaysF_{\intervco{0}{m}}\left(  \varphi \until \left(\varphi \wedge \next \psi\right)\right)$.
 
For the converse direction, from $(\tuple{\H,\T},\tmf), i \models \alwaysF_{\intervco{0}{m}} \left( \varphi \until \left(\varphi \wedge \next \psi\right)\right)$ it follows that there exists $j > i$ such that $\tau(j)-\tau(i) \ge m$, $(\tuple{\H,\T},\tmf), j \models \psi$ and $(\tuple{\H,\T},\tmf), k \models \varphi$ for all $k\in \intervco{i}{j}$.
Since $(\tuple{\H,\T},\tmf), i \models  \eventuallyF_{\intervcc{m}{n}} \psi$ there exists $j'> i$ such that $\tau(j') - \tau(i) \ge m$, $\tau(j') - \tau(i) \leq n$ and $(\tuple{\H,\T},\tmf),j' \models \psi$.

If $j' < j$ we can easily conclude that $(\tuple{\H,\T},\tmf), i \models \varphi \until_{\intervcc{m}{n}} \psi$.
If $j' \ge j$ then $\tau(j') \ge \tau(j)$. 
Since $\tau(j') - \tau(i) \leq n $ and $\tau(j') > \tau(j)$, $\tau(j)-\tau(i) \leq n$ so $\tau(j) - \tau(i) \in \intervcc{m}{n}$, which leads to $(\tuple{\H,\T},\tmf),i \models \varphi \until_{\intervcc{m}{n}} \psi$.\\ 

\noindent For the third equivalence $\varphi \until_{\intervoo{m}{n}} \psi \equiv \eventuallyF_{\intervoo{m}{n}} \psi \wedge \alwaysF_{\intervcc{0}{m}} \left( \varphi \until \left(\varphi \wedge \next \psi\right)\right)$,\\
from left to right, if $(\tuple{\H,\T},\tmf), i \models \varphi \until_{\intervoo{m}{n}} \psi$ then there exists $j \ge i$ such that $\rangeoo{\tau(j)-\tau(i)}{m}{n}$, $(\tuple{\H,\T},\tmf), j \models \psi$ and for all $\rangeco{k}{i}{j}$, $(\tuple{\H,\T},\tmf), k \models \varphi$.
From  $\rangeoo{\tau(j)-\tau(i)}{m}{n}$, $j\ge i$ and $(\tuple{\H,\T},\tmf), j \models \psi$ it follows that $(\tuple{\H,\T},\tmf), i \models \eventuallyF_{\intervoo{m}{n}} \psi$.
Moreover, since $m \not = 0$, $\tau(j) - \tau(i) \not= 0$ so $j \not = i$, which implies that $j > i$. 
As a consequence $(\tuple{\H,\T},\tmf), j-1 \models \varphi \wedge \next \psi$ and $(\tuple{\H,\T},\tmf), t \models \varphi$ for all $i \le t < j-1$.

Take any arbitrary $y \ge i$. 
If $y \ge j$ then $\tau(y) - \tau(i) > m$ because $\tau(y) \ge \tau(j)$ and $\tau(j)-\tau(i) > m$. Therefore, $\tau(y) - \tau(i) \not \in \intervcc{0}{m}$.
If $y < j$ then $(\tuple{\H,\T},\tmf),y \models \varphi \until \left(\varphi \wedge \next \psi\right)$.
Since $y$ was arbitrary chosen, it follows that $(\tuple{\H,\T},\tmf), i \models \alwaysF_{\intervcc{0}{m}}\left(  \varphi \until \left(\varphi \wedge \next \psi\right)\right)$.
 
For the converse direction, from $(\tuple{\H,\T},\tmf), i \models \alwaysF_{\intervcc{0}{m}} \left( \varphi \until \left(\varphi \wedge \next \psi\right)\right)$ it follows that there exists $j > i$ such that $\tau(j)-\tau(i) > m$, $(\tuple{\H,\T},\tmf), j \models \psi$ and $(\tuple{\H,\T},\tmf), k \models \varphi$ for all $k\in \intervco{i}{j}$.
Since $(\tuple{\H,\T},\tmf), i \models  \eventuallyF_{\intervoo{m}{n}} \psi$ there exists $j'> i$ such that $\tau(j') - \tau(i) > m$, $\tau(j') - \tau(i) < n$ and $(\tuple{\H,\T},\tmf),j' \models \psi$.

If $j' < j$ we can easily conclude that $(\tuple{\H,\T},\tmf), i \models \varphi \until_{\intervoo{m}{n}} \psi$.
If $j' \ge j$ then $\tau(j') \ge \tau(j)$. 
Since $\tau(j') - \tau(i) < n $ and $\tau(j') > \tau(j)$, $\tau(j)-\tau(i) < n$, so $\tau(j) - \tau(i) \in \intervoo{m}{n}$, which leads to $(\tuple{\H,\T},\tmf),i \models \varphi \until_{\intervoo{m}{n}} \psi$.\\ 

\noindent For the fourth equivalence $\varphi \until_{\intervoc{m}{n}} \psi \equiv \eventuallyF_{\intervoc{m}{n}} \psi \wedge \alwaysF_{\intervcc{0}{m}} \left( \varphi \until \left(\varphi \wedge \next \psi\right)\right)$,\\
from left to right, if $(\tuple{\H,\T},\tmf), i \models \varphi \until_{\intervoc{m}{n}} \psi$ then there exists $j \ge i$ such that $\rangeoc{\tau(j)-\tau(i)}{m}{n}$, $(\tuple{\H,\T},\tmf), j \models \psi$ and for all $\rangeco{k}{i}{j}$, $(\tuple{\H,\T},\tmf), k \models \varphi$.
From  $\rangeoc{\tau(j)-\tau(i)}{m}{n}$, $j\ge i$ and $(\tuple{\H,\T},\tmf), j \models \psi$ it follows that $(\tuple{\H,\T},\tmf), i \models \eventuallyF_{\intervoc{m}{n}} \psi$.
Moreover, since $m \not = 0$, $\tau(j) - \tau(i) \not= 0$ so $j \not = i$, which implies that $j > i$. 
As a consequence $(\tuple{\H,\T},\tmf), j-1 \models \varphi \wedge \next \psi$ and $(\tuple{\H,\T},\tmf), t \models \varphi$ for all $i \le t < j-1$.

Take any arbitrary $y \ge i$. 
If $y \ge j$ then $\tau(y) - \tau(i) > m$ because $\tau(y) \ge \tau(j)$ and $\tau(j)-\tau(i) > m$. Therefore, $\tau(y) - \tau(i) \not \in \intervcc{0}{m}$.
If $y < j$ then $(\tuple{\H,\T},\tmf),y \models \varphi \until \left(\varphi \wedge \next \psi\right)$.
Since $y$ was arbitrary chosen, it follows that $(\tuple{\H,\T},\tmf), i \models \alwaysF_{\intervcc{0}{m}}\left(  \varphi \until \left(\varphi \wedge \next \psi\right)\right)$.
 
For the converse direction, from $(\tuple{\H,\T},\tmf), i \models \alwaysF_{\intervcc{0}{m}} \left( \varphi \until \left(\varphi \wedge \next \psi\right)\right)$ it follows that there exists $j > i$ such that $\tau(j)-\tau(i) > m$, $(\tuple{\H,\T},\tmf), j \models \psi$ and $(\tuple{\H,\T},\tmf), k \models \varphi$ for all $k\in \intervco{i}{j}$.
Since $(\tuple{\H,\T},\tmf), i \models  \eventuallyF_{\intervoc{m}{n}} \psi$ there exists $j'> i$ such that $\tau(j') - \tau(i) > m$, $\tau(j') - \tau(i) \leq n$ and $(\tuple{\H,\T},\tmf),j' \models \psi$.

If $j' < j$ we can easily conclude that $(\tuple{\H,\T},\tmf), i \models \varphi \until_{\intervoc{m}{n}} \psi$.
If $j' \ge j$ then $\tau(j') \ge \tau(j)$. 
Since $\tau(j') - \tau(i) \leq n $ and $\tau(j') > \tau(j)$, $\tau(j)-\tau(i) \leq n$, so $\tau(j) - \tau(i) \in \intervoc{m}{n}$, which leads to $(\tuple{\H,\T},\tmf),i \models \varphi \until_{\intervoc{m}{n}} \psi$.\\

\noindent For the fifth equivalence $\varphi \until_{\intervco{0}{n}} \psi \equiv \eventuallyF_{\intervco{0}{n}}\psi \wedge \varphi \until \psi$,\\
from left to right, if $(\tuple{\H,\T},\tmf), i \models \varphi \until_{\intervco{0}{n}} \psi$ then there exists $j \ge i$ such that $\rangeco{\tau(j)-\tau(i)}{0}{n}$, $(\tuple{\H,\T},\tmf), j \models \psi$ and for all $\rangeco{k}{i}{j}$, $(\tuple{\H,\T},\tmf), k \models \varphi$.
This already implies $(\tuple{\H,\T},\tmf), i \models \eventuallyF_{\intervco{0}{n}}\psi$. 
Furthermore, since $\intervco{0}{n} \subseteq \intervco{0}{\omega}$, we can also derive $(\tuple{\H,\T},\tmf), i \models \varphi \until \psi$ from $(\tuple{\H,\T},\tmf), i \models \varphi \until_{\intervco{0}{n}} \psi$.
Putting both implications together we get $(\tuple{\H,\T},\tmf), i \models\eventuallyF_{\intervco{0}{n}}\psi \wedge \varphi \until \psi$.
 
For the converse direction, from $(\tuple{\H,\T},\tmf), i \models \eventuallyF_{\intervco{0}{n}}\psi \wedge \varphi \until \psi$, we can derive $(\tuple{\H,\T},\tmf), i \models \varphi \until \psi$ and therefore there exists $j \ge i$ s.t. $(\tuple{\H,\T},\tmf), j \models \psi$ and $\rangeco{\tau(j)-\tau(i)}{0}{\omega}$ and $(\tuple{\H,\T},\tmf), k \models \varphi$ for all $\rangeco{k}{i}{j}$. 
$(\tuple{\H,\T},\tmf), i \models  \eventuallyF_{\intervco{0}{n}} \psi$ implies $(\tuple{\H,\T},\tmf), j' \models \psi$ for some $j' \ge i$ with $\rangeco{\tau(j')-\tau(i)}{0}{n}$.
Now there are two cases to consider. 
If $j' < j$ we can easily conclude that $(\tuple{\H,\T},\tmf), i \models \varphi \until_{\intervco{0}{n}} \psi$.
If $j' \ge j$ then $\tau(j) - \tau(i) \in \intervco{0}{n}$ and therefore $(\tuple{\H,\T},\tmf), i \models \varphi \until_{\intervco{0}{n}} \psi$.\\

\noindent For the sixth equivalence $\varphi \until_{\intervcc{0}{n}} \psi \equiv  \eventuallyF_{\intervcc{0}{n}}\psi \wedge \varphi \until \psi$,\\
from left to right, if $(\tuple{\H,\T},\tmf), i \models \varphi \until_{\intervcc{0}{n}} \psi$ then there exists $j \ge i$ such that $\rangecc{\tau(j)-\tau(i)}{0}{n}$, $(\tuple{\H,\T},\tmf), j \models \psi$ and for all $\rangeco{k}{i}{j}$, $(\tuple{\H,\T},\tmf), k \models \varphi$.
This already implies $(\tuple{\H,\T},\tmf), i \models \eventuallyF_{\intervcc{0}{n}}\psi$. 
Furthermore, since $\intervcc{0}{n} \subseteq \intervco{0}{\omega}$, we can also derive $(\tuple{\H,\T},\tmf), i \models \varphi \until \psi$ from $(\tuple{\H,\T},\tmf), i \models \varphi \until_{\intervcc{0}{n}} \psi$.
Putting both implications together we get $(\tuple{\H,\T},\tmf), i \models\eventuallyF_{\intervco{0}{n}}\psi \wedge \varphi \until \psi$.
 
For the converse direction, from $(\tuple{\H,\T},\tmf), i \models \eventuallyF_{\intervcc{0}{n}}\psi \wedge \varphi \until \psi$, we can derive $(\tuple{\H,\T},\tmf), i \models \varphi \until \psi$ and therefore there exists $j \ge i$ s.t. $(\tuple{\H,\T},\tmf), j \models \psi$ and $\rangeco{\tau(j)-\tau(i)}{0}{\omega}$ and $(\tuple{\H,\T},\tmf), k \models \varphi$ for all $\rangeco{k}{i}{j}$. 
$(\tuple{\H,\T},\tmf), i \models  \eventuallyF_{\intervcc{0}{n}} \psi$ implies $(\tuple{\H,\T},\tmf), j' \models \psi$ for some $j' \ge i$ with $\rangecc{\tau(j')-\tau(i)}{0}{n}$.
Now there are two cases to consider. 
If $j' < j$ we can easily conclude that $(\tuple{\H,\T},\tmf), i \models \varphi \until_{\intervcc{0}{n}} \psi$.
If $j' \ge j$ then $\tau(j) - \tau(i) \in \intervcc{0}{n}$ and therefore $(\tuple{\H,\T},\tmf), i \models \varphi \until_{\intervcc{0}{n}} \psi$.\\

\noindent For the seventh equivalence $\varphi \until_{\intervoo{0}{n}} \psi \equiv  \eventuallyF_{\intervoo{0}{n}}\psi \wedge \varphi \until \left(\varphi \wedge \next \psi\right)$,\\
from left to right, if $(\tuple{\H,\T},\tmf), i \models \varphi \until_{\intervoo{0}{n}} \psi$ then there exists $j > i$ such that $\rangeoo{\tau(j)-\tau(i)}{0}{n}$, $(\tuple{\H,\T},\tmf), j \models \psi$ and for all $\rangeco{k}{i}{j}$, $(\tuple{\H,\T},\tmf), k \models \varphi$.
This already implies $(\tuple{\H,\T},\tmf), i \models \eventuallyF_{\intervoo{0}{n}}\psi$. 
Furthermore, since $j > i$, we know\\ $(\tuple{\H,\T},\tmf), j-1 \models \varphi \wedge \next \psi$ and $(\tuple{\H,\T},\tmf), t \models \varphi$ for all $ i \leq t < i-1$ and therefore $(\tuple{\H,\T},\tmf), i \models \eventuallyF_{\intervoo{0}{n}}\psi \wedge \varphi \until \left(\varphi \wedge \next \psi\right)$.
 
For the converse direction, $(\tuple{\H,\T},\tmf), i \models \varphi \until \left(\varphi \wedge \next \psi\right)$ implies that there is $j >i$ s.t. $(\tuple{\H,\T},\tmf),j \models \psi$ and $ (\tuple{\H,\T},\tmf), k \models \varphi$ for all $\rangeco{k}{i}{j}$ which implies $(\tuple{\H,\T},\tmf), i \models \varphi \until_{>0} \psi$. 
Furthermore, since $(\tuple{\H,\T},\tmf), i \models \eventuallyF_{\intervoo{0}{n}}\psi$ there exists $j' > j$ s.t. $\tau(j')- \tau(j) \in \intervoo{0}{n}$ and $(\tuple{\H,\T},\tmf), j' \models \psi$. 
This together with $(\tuple{\H,\T},\tmf), i \models \varphi \until_{>0} \psi$ implies $(\tuple{\H,\T},\tmf), i \models \varphi \until_{\intervoo{0}{n}} \psi$ by following similar reasoning as in the previous cases.\\

\noindent For the eighth equivalence $\varphi \until_{\intervoc{0}{n}} \psi \equiv  \eventuallyF_{\intervoc{0}{n}}\psi \wedge \varphi \until \left(\varphi \wedge \next \psi\right)$,\\
from left to right, if $(\tuple{\H,\T},\tmf), i \models \varphi \until_{\intervoc{0}{n}} \psi$ then there exists $j > i$ such that $\rangeoc{\tau(j)-\tau(i)}{0}{n}$, $(\tuple{\H,\T},\tmf), j \models \psi$ and for all $\rangeco{k}{i}{j}$, $(\tuple{\H,\T},\tmf), k \models \varphi$.
This already implies $(\tuple{\H,\T},\tmf), i \models \eventuallyF_{\intervoc{0}{n}}\psi$. 
Furthermore, since $j > i$, we know\\ $(\tuple{\H,\T},\tmf), j-1 \models \varphi \wedge \next \psi$ and $(\tuple{\H,\T},\tmf), t \models \varphi$ for all $ i \leq t < i-1$ and therefore $(\tuple{\H,\T},\tmf), i \models \eventuallyF_{\intervoo{0}{n}}\psi \wedge \varphi \until \left(\varphi \wedge \next \psi\right)$.
 
For the converse direction, $(\tuple{\H,\T},\tmf), i \models \varphi \until \left(\varphi \wedge \next \psi\right)$ implies that there is $j >i$ s.t. $(\tuple{\H,\T},\tmf),j \models \psi$ and $ (\tuple{\H,\T},\tmf), k \models \varphi$ for all $\rangeco{k}{i}{j}$ which implies $(\tuple{\H,\T},\tmf), i \models \varphi \until_{>0} \psi$. 
Furthermore, since $(\tuple{\H,\T},\tmf), i \models \eventuallyF_{\intervoc{0}{n}}\psi$ there exists $j' > j$ s.t. $\tau(j')- \tau(j) \in \intervoc{0}{n}$ and $(\tuple{\H,\T},\tmf), j' \models \psi$. 
This together with $(\tuple{\H,\T},\tmf), i \models \varphi \until_{>0} \psi$ implies $(\tuple{\H,\T},\tmf), i \models \varphi \until_{\intervoc{0}{n}} \psi$ by following similar reasoning as in the previous cases.\\

\noindent The case for Release can be proven by applying Corollary \ref{BDT} (Boolean Duality) and uniform substitution to the respective Until cases. The cases for Since and Trigger follow from applying Theorem \ref{th:temporal_duality} (Temporal Duality) to the Until and Release cases respectively.  

\end{proofof}

\begin{proofof}{Theorem~\ref{prop:sequivalence}}
	From right to left, if $\Gamma_1$ and $\Gamma_2$ are \MHT{} equivalent then $\Gamma_1$ and $\Gamma_2$ have the same \MHT{} models. 
	As a consequence, $\Gamma_1\cup \Gamma$ and $\Gamma_2 \cup \Gamma$ have the same \MHT{} models.
	Therefore, $\Gamma_1\cup \Gamma$ and $\Gamma_2 \cup \Gamma$ have the same \MEL{} models.
	Since $\Gamma$ is any arbitrary temporal theory, $\Gamma_1$ and $\Gamma_2$ are strongly equivalent.
	
	For the converse direction let us assume that $\Gamma_1$ and $\Gamma_2$ are strongly equivalent but they are not \MHT{} equivalent.
	We consider two cases:
	
	\begin{enumerate}
		\item $\Gamma_1$ and $\Gamma_2$ are not \MTL{} equivalent.
		Assume, without loss of generality, that there exists a total \MHT{} model $(\tuple{\T,\T}, \tmf)$ such that $(\tuple{\T,\T}, \tmf),0 \models \Gamma_1$ but $(\tuple{\T,\T}, \tmf), 0 \not \models \Gamma_2$.
		Since $(\tuple{\T,\T}, \tmf)$ is total, it follows that $(\tuple{\T,\T}, \tmf), 0 \models \Gamma_1 \cup \EM{\PV}$ and $(\tuple{\T,\T}, \tmf), 0\not \models \Gamma_2 \cup \EM{\PV}$.
		Moreover, $(\tuple{\T,\T}, \tmf)$ is an equilibrium model of $\Gamma_1 \cup \EM{\PV}$ (since for any $\H < \T$ , $(\tuple{\H,\T}, \tmf), 0 \not \models\EM{\PV}$) but not of $\Gamma_2 \cup \EM{\PV}$. 
		\item $\Gamma_1$ and $\Gamma_2$ are \MTL{} equivalent.
		Therefore, without loss of generality, there exists a \MHT{} interpretation $(\tuple{\H,\T}, \tmf)$ with $\H < \T$ such that 
		\begin{enumerate}
			\item $(\tuple{\T,\T}, \tmf),0 \models \Gamma_1$ iff $(\tuple{\T,\T}, \tmf),0 \models \Gamma_2$ because $\Gamma_1$ and $\Gamma_2$ are \MTL{} equivalent.
			\item $(\tuple{\H,\T}, \tmf),0 \models \Gamma_1$ and $(\tuple{\H,\T}, \tmf),0 \not \models \Gamma_2$ because $\Gamma_1$ and $\Gamma_2$ are not \MHT{} equivalent.
		\end{enumerate}
	\end{enumerate}
	
	Since  $(\tuple{\H,\T}, \tmf),0 \not \models \Gamma_2$, there exists $\varphi \in \Gamma_2$ such that $(\tuple{\H,\T}, \tmf),0 \not \models \varphi$.	
	Moreover, since $(\tuple{\H,\T}, \tmf),0 \models \Gamma_1$ then $(\tuple{\T,\T}, \tmf),0 \models \Gamma_1$ so $(\tuple{\T,\T}, \tmf),0 \models \Gamma_2$ and so $(\tuple{\T,\T}, \tmf),0 \models \varphi$.
	
	Let us consider the theory $\Gamma \eqdef \lbrace  \varphi \to \psi \mid \psi \in \EM{\PV}\rbrace$.
	Since $(\tuple{\H,\T}, \tmf), 0 \not \models \varphi$ and $(\tuple{\T,\T}, \tmf),0 \models \EM{\PV}$ then $(\tuple{\H,\T}, \tmf),0 \models \Gamma$.
	Therefore, $(\tuple{\H,\T}, \tmf), 0 \models \Gamma_1 \cup \Gamma$ so $(\tuple{\T,\T}, \tmf)$ is not an equilibrium model of $\Gamma_1 \cup \Gamma$. 
	Since $\Gamma_1$ and $\Gamma_2$ are strongly equivalent, $(\tuple{\T,\T}, \tmf)$ cannot be an equilibrium model of $\Gamma_2 \cup \Gamma$. 
	Since $(\tuple{\T,\T}, \tmf),0 \models \Gamma_2$ and $(\tuple{\T,\T}, \tmf), 0 \models \Gamma$ then the minimality condition must fail. 
	This means that there must exist $\H'< \T$ such that $(\tuple{\H',\T}, \tmf), 0 \models \Gamma_2 \cup \Gamma$. 
	Since $(\tuple{\H',\T}, \tmf), 0 \models \Gamma_2$ then $(\tuple{\H',\T}, \tmf), 0 \models \varphi$.
	Since  $(\tuple{\H',\T}, \tmf), 0 \models \varphi$ and $(\tuple{\H',\T}, \tmf), 0 \models \Gamma$ then  $(\tuple{\H',\T}, \tmf), 0 \models \EM{\PV}$, which contradicts Proposition~\ref{prop:total}.						
\end{proofof}

\begin{proofof}{Theorem~\ref{thm:kamp}}
The proof goes by structural induction.

\paragraph{Base case} let us consider first the case of a propositional variable $p$.
From left to right, if $(\tuple{\H,\T}, \tau), i \models \varphi$ then $p \in H_i$.
By definition, $\tau(i) \in D$ and $p(\tau(i)) \in H$.
Therefore $\tuple{(D,\sigma),H,T} \models \tr{p}_{\tau_{i}}$.

\paragraph{Inductive case: propositional connectives}

\begin{itemize}
	\item Case $\varphi \wedge \psi$: 
		\begin{eqnarray*}
	(\tuple{\H,\T}, \tmf), i \models \varphi \wedge \psi &\text{iff}&	(\tuple{\H,\T}, \tmf), i \models \varphi \text{ and } (\tuple{\H,\T}, \tmf), i \models \psi \\
		& \stackrel{IH}{\text{iff}}& \tuple{(D,\sigma),H,T} \models \tr{\varphi}_{\tau_{i}} \text{ and }\tuple{(D,\sigma),H,T} \models \tr{\psi}_{\tau_{i}} \\
		&\text{ iff }& \tuple{(D,\sigma),H,T} \models \tr{\varphi\wedge \psi}_{\tau_{i}}.
		\end{eqnarray*}


	\item Case $\varphi \vee \psi$:
	
	\begin{eqnarray*}
		(\tuple{\H,\T}, \tmf), i \models \varphi \vee \psi &\text{iff}&	(\tuple{\H,\T}, \tmf), i \models \varphi \text{ or } (\tuple{\H,\T}, \tmf), i \models \psi \\
		& \stackrel{IH}{\text{iff}}& \tuple{(D,\sigma),H,T} \models \tr{\varphi}_{\tau_{i}} \text{ or }\tuple{(D,\sigma),H,T} \models \tr{\psi}_{\tau_{i}} \\
		&\text{ iff }& \tuple{(D,\sigma),H,T} \models \tr{\varphi\vee \psi}_{\tau_{i}}.
	\end{eqnarray*}

%
	\item Case $\varphi \rightarrow \psi$:
In the first case, $(\tuple{\H,\T}, \tmf), i \models \varphi \to \psi$ iff for all $\otimes \in \lbrace \H,\T \rbrace$, either $\tuple{\otimes,\T,\tau},i \not \models \varphi$ or $\tuple{\otimes,\T,\tau},i \models \psi$. 
Since both $(\tuple{\H,\T}, \tmf)$ and $(\tuple{\T,\T}, \tmf)$ are models, we can apply the induction hypothesis~\eqref{thm:kamp:ih1} on both so we get iff for all $\oplus\in \lbrace H,T\rbrace$, either $\tuple{(D,\sigma),\oplus,T} \not \models \tr{\varphi}_{\tau_{i}}$ or $\tuple{(D,\sigma),\oplus,T} \models \tr{\psi}_{\tau_{i}}$.
Therefore, $\tuple{(D,\sigma),H,T} \models \tr{\varphi\to \psi}_{\tau_{i}}$.

	
\end{itemize}

\paragraph{Inductive case: metric temporal operators} For simplicity we will consider intervals of the form $[m,n)$ where $n \not = \omega$.

\begin{itemize}
	\item Case $\metricol{\next}{m}{n}\varphi$: if $(\tuple{\H,\T}, \tmf),i \models \metricol{\next}{m}{n}\varphi$ then there exists $i+1 < \lambda$ such that $m \le \tau(i+1)-\tau(i) < n$ and $(\tuple{\H,\T}, \tmf), i+1 \models \varphi$.
	By the induction hypothesis~\eqref{thm:kamp:ih1}
	we get $\tuple{(D,\sigma),H,T} \models \tr{\varphi}_{\tau_{i+1}}$
	From $m \le \tau(i+1)-\tau(i) < n$ we conclude that $\tau(i) \peq{-m} \tau(i+1) \pt{n} \tau(i)$.
	Since we are dealing with strict traces, $\tau(i) < \tau(i+1)$ and, moreover, there is no other $\tau(j)$ in between.
	Therefore, for all $d \in D$, not $\tau(i) < d < \tau(i+1)$, so $\tuple{(D,\sigma),H,T} \models \neg \exists z\; \tau(i) < z < \tau(i+1)$;
	Since $\tau(i+1)\in D$, we conclude that there exists $d \in D$ such that 
	
	\begin{equation*}
	\tuple{(D,\sigma),H,T} \models  \tau(i) < d \wedge \left( \neg \exists z\; \tau(i) < z < d\right) \wedge \tau(i) \peq{-m} d \pt{n} \tau(i) \wedge \tr{\varphi}_{d}.
	\end{equation*}
	
	Therefore, 
	
	\begin{equation*}
		\tuple{(D,\sigma),H,T} \models \exists y\; \left( \tau(i) < y \wedge \left( \neg \exists z\; \tau(i) < z < y\right) \wedge \tau(i) \peq{-m} y \pt{n} \tau(i) \wedge \tr{\varphi}_{y} \right).
	\end{equation*}
	From this we conclude 	$\tuple{(D,\sigma),H,T} \models \tr{\metricol{\next}{m}{n}\varphi}_{\tau(i)}$.
	
	Conversely, if $\tuple{(D,\sigma),H,T} \models \tr{\metricol{\next}{m}{n}\varphi}_{\tau(i)}$ then, by definition, 	
	\begin{equation*}
		\tuple{(D,\sigma),H,T} \models \exists y\; \left( \tau(i) < y \wedge \left( \neg \exists z\; \tau(i) < z < y\right) \wedge \tau(i) \peq{-m} y \pt{n} \tau(i) \wedge \tr{\varphi}_{y} \right), 
	\end{equation*} 
	Therefore, there exists $d \in D$ such that 	

	\begin{equation*}
		\tuple{(D,\sigma),H,T} \models \tau(i) < d \wedge \left( \neg \exists z\; \tau(i) < z < d\right) \wedge \tau(i) \peq{-m} d \pt{n} \tau(i) \wedge \tr{\varphi}_{d}.
	\end{equation*} 
	From $\tuple{(D,\sigma),H,T} \models \tau(i) < d \wedge \left( \neg \exists z\; \tau(i) < z < d\right)$ and the construction of $\tuple{(D,\sigma),H,T}$ we conclude that $d = \tau(i)$.
	Since $\tuple{(D,\sigma),H,T} \models \tau(i) \peq{-m} \tau(i+1) \pt{n} \tau(i)$, we conclude that $m \le \tau(i+1) - \tau(i) < n$.
	By the induction hypothesis~\eqref{thm:kamp:ih1}, $(\tuple{\H,\T}, \tmf),i+1 \models \varphi$. 
	Therefore, $(\tuple{\H,\T}, \tmf), i \models \metricol{\next}{m}{n} \varphi$.
	

	\item Case $\metricol{\wnext}{m}{n}\varphi$: in the first case, if $(\tuple{\H,\T}, \tmf),i \not \models \metricol{\wnext}{m}{n}$ then $i+1 < \lambda$ such that $m \le \tau(i+1)-\tau(i) < n$ and $(\tuple{\H,\T}, \tmf), i+1 \not\models \varphi$.
	By the induction hypothesis~\eqref{thm:kamp:ih1} we get $\tuple{(D,\sigma),H,T} \not \models \tr{\varphi}_{\tau_{i+1}}$
	From $m \le \tau(i+1)-\tau(i) < n$ we conclude that $\tau(i) \peq{-m} \tau(i+1) \pt{n} \tau(i)$.
	Since we are dealing with strict traces, $\tau(i) < \tau(i+1)$ and, moreover, there is no other $\tau(j)$ in between.
	Therefore, for all $d \in D$, not $\tau(i) < d < \tau(i+1)$, so $\tuple{(D,\sigma),H,T} \models \neg \exists z\; \tau(i) < z < \tau(i+1)$;
	Since $\tau(i+1)\in D$, we conclude that there exists $d \in D$ such that 
	
	\begin{equation*}
		\tuple{(D,\sigma),H,T} \not \models  \left(\tau(i) < d \wedge \left( \neg \exists z\; \tau(i) < z < d\right) \wedge \tau(i) \peq{-m} d \pt{n} \tau(i)\right) \to \tr{\varphi}_{d}.
	\end{equation*}
	
	Therefore, 
	
	\begin{equation*}
		\tuple{(D,\sigma),H,T} \not \models \forall y\; \left( \left(\tau(i) < y \wedge \left( \neg \exists z\; \tau(i) < z < y\right) \wedge \tau(i) \peq{-m} y \pt{n} \tau(i)\right) \to \tr{\varphi}_{y} \right).
	\end{equation*}
	From this we conclude 	$\tuple{(D,\sigma),H,T} \not \models \tr{\metricol{\wnext}{m}{n}\varphi}_{\tau(i)}$.
	
	Conversely, if $\tuple{(D,\sigma),H,T} \not \models \tr{\metricol{\wnext}{m}{n}\varphi}_{\tau(i)}$ then, by definition, 	
	\begin{equation*}
		\tuple{(D,\sigma),H,T} \not \models \forall y\; \left( \left(\tau(i) < y \wedge \left( \neg \exists z\; \tau(i) < z < y\right) \wedge \tau(i) \peq{-m} y \pt{n} \tau(i)\right) \to \tr{\varphi}_{y} \right), 
	\end{equation*} 
	Therefore, there exists $d \in D$ such that 	
	
	\begin{equation*}
		\tuple{(D,\sigma),H,T} \not \models \left(\tau(i) < d \wedge \left( \neg \exists z\; \tau(i) < z < d\right) \wedge \tau(i) \peq{-m} d \pt{n} \tau(i)\right) \to \tr{\varphi}_{d}.
	\end{equation*} 

	We consider two cases: 
	\begin{enumerate}
		\item $\tuple{(D,\sigma),H,T} \models \left(\tau(i) < d \wedge \left( \neg \exists z\; \tau(i) < z < d\right) \wedge \tau(i) \peq{-m} d \pt{n} \tau(i)\right)$ and $\tuple{(D,\sigma),H,T} \not \models \tr{\varphi}_{d}$;
		\item $\tuple{(D,\sigma),T,T} \models \left(\tau(i) < d \wedge \left( \neg \exists z\; \tau(i) < z < d\right) \wedge \tau(i) \peq{-m} d \pt{n} \tau(i)\right)$ and $\tuple{(D,\sigma),T,T} \not \models \tr{\varphi}_{d}$;
	\end{enumerate}

	In any of the previous cases, we conclude that $d=\tau(i+1)$, $m \le \tau(i+1) - \tau(i) < n$ and $\tuple{(D,\sigma),H,T} \not \models \tr{\varphi}_{\tau(i)}$.

	By the induction hypothesis~\eqref{thm:kamp:ih1}, $(\tuple{\H,\T}, \tmf),i+1 \not \models \varphi$. 
	Therefore, $(\tuple{\H,\T}, \tmf), i \not \models \metricol{\wnext}{m}{n} \varphi$.
	

	\item Case $\varphi  \metricol{\until}{m}{n}\psi$: for the first item, if $(\tuple{\H,\T}, \tmf),i \models \varphi \metricol{\until}{m}{n}\psi$ then there exists $j \ge i$ such that $m \le \tau(j)-\tau(i) < n$,  $(\tuple{\H,\T}, \tmf),j \models \psi$ and for all $i \le k < j$, $(\tuple{\H,\T}, \tmf),k \models \varphi$.
	Since  $m \le \tau(j)-\tau(i) < n$ then $\tuple{(D,\sigma),H,T} \models \tau(i) \le \tau(j) \wedge \tau(i) \peq{-m} \tau(j)  \pt{n} \tau(i)$.
	From $(\tuple{\H,\T}, \tmf),j \models \psi$ and the induction hypothesis~\eqref{thm:kamp:ih1} we get $\tuple{(D,\sigma),H,T} \models \tr{\psi}_{\tau(j)}$.
	From  $(\tuple{\H,\T}, \tmf),k \models \varphi$ for all $i \le k < j$ and the induction hypothesis~\eqref{thm:kamp:ih1} we get $\tuple{(D,\sigma),H,T} \models \tr{\varphi}_{d}$ for all $d \in \lbrace\tau(k) \mid \tau(i) \le k < \tau(j)\rbrace$.
	By the semantics, $\tuple{(D,\sigma),H,T} \models \forall z \left(\tau(i) \le z < \tau(j) \to \tr{\varphi}_z\right) $.
	Therefore, $\tuple{(D,\sigma),H,T} \models \tr{\varphi \metricol{\until}{m}{n}\psi}_{\tau(i)}$.
	Conversely, if $\tuple{(D,\sigma),H,T} \models \tr{\varphi \metricol{\until}{m}{n}\psi}_{\tau(i)}$ then 
	
	\begin{equation*}
	\tuple{(D,\sigma),H,T} \models \exists y\; \left(\tau(i) \le y \wedge \tau(i) \peq{-m} y  \pt{n} \tau(i) \wedge
	\tr{\psi}_y  \wedge \forall z \left(\tau(i) \le z < y  \rightarrow
	\tr{\varphi}_z\right)\right).
	\end{equation*}

	This means that there exists $\tau(j) \in D$ such that 	$\tuple{(D,\sigma),H,T} \models \left(\tau(i) \le \tau(j) \wedge \tau(i) \peq{-m} \tau(j)\right)$, $\tuple{(D,\sigma),H,T} \models\tr{\psi}_{\tau(j)}$ and $\tuple{(D,\sigma),H,T} \models  \forall z \left(\tau(i) \le z < \tau(j)  \rightarrow
	\tr{\varphi}_z\right)$.
	From $\tuple{(D,\sigma),H,T} \models \left(\tau(i) \le \tau(j) \wedge \tau(i) \peq{-m} \tau(j)\right)$ it follows that $i \le j$ and $\tau(j)-\tau(i) \in [m,n)$.
	By induction, $(\tuple{\H,\T}, \tmf), j \models \psi$.
	From $\tuple{(D,\sigma),H,T} \models  \forall z \left(\tau(i) \le z < \tau(j)  \rightarrow
	\tr{\varphi}_z\right)$ it follows that for all $\tau(k) \in D$, if $\tau(i) \le \tau(k) < \tau(j)$ then $\tuple{(D,\sigma),H,T} \models \tr{\varphi}_{\tau(k)}$.
	By induction we get that $(\tuple{\H,\T}, \tmf),k \models \varphi$, for all $i \le k < j$. 
	From all previous statements it follows $(\tuple{\H,\T}, \tmf), i \models  \varphi  \metricol{\until}{m}{n}\psi$.
	
	\item Case $\varphi\metricol{\release}{m}{n}\psi$: from left to right, assume by contraposition that $\tuple{(D,\sigma),H,T} \not \models \tr{\varphi \metricol{\until}{m}{n}\psi}_{\tau(i)}$ then 
	
	\begin{equation*}
		\tuple{(D,\sigma),H,T} \not \models \forall y\; \left(\left(\tau(i) \le y \wedge \tau(i) \peq{-m} y  \pt{n}
		\tau(i)\right) \rightarrow \left( \tr{\psi}_y  \vee \exists z \left( \tau(i)
		\le z < y \wedge \tr{\varphi}_z\right)\right)\right).
	\end{equation*}
	
	Therefore, there exists $\tau(j) \in D$ such that 
	
	\begin{equation*}
		\tuple{(D,\sigma),H,T} \not \models \left(\tau(i) \le \tau(j) \wedge \tau(i) \peq{-m} \tau(j)  \pt{n}
		\tau(i)\right) \rightarrow \left( \tr{\psi}_{\tau(j)}  \vee \exists z \left( \tau(i)
		\le z < y \wedge \tr{\varphi}_z\right)\right).
	\end{equation*}

	From this and a some \HT{} reasoning\footnote{Using persistency and the fact that the satisfaction of the expression $(\tau(i) \le \tau(j)) \wedge (\tau(i) \peq{-m} \tau(j)  \pt{n} \tau(i))	$  is not \HT{}-dependent.} we can conclude that $\tuple{(D,\sigma),H,T} \models \left(\tau(i) \le \tau(j) \wedge \tau(i) \peq{-m} \tau(j)  \pt{n}
	\tau(i)\right)$ but  $\tuple{(D,\sigma),H,T} \not \models  \tr{\psi}_{\tau(j)}$ and $\tuple{(D,\sigma),H,T} \not \models\exists z \left( \tau(i)	\le z < y \wedge \tr{\varphi}_z\right)$.	
	From $\tuple{(D,\sigma),H,T} \models \left(\tau(i) \le \tau(j) \wedge \tau(i) \peq{-m} \tau(j)\right)$ it follows that $i \le j$ and $\tau(j)-\tau(i) \in [m,n)$.	
	By induction~\eqref{thm:kamp:ih1}, $(\tuple{\H,\T}, \tmf), j \not \models \psi$.
	From $\tuple{(D,\sigma),H,T} \not \models  \exists  z \left(\tau(i) \le z < \tau(j)  \wedge
	\tr{\varphi}_z\right)$ it follows that for all $\tau(k) \in D$, if $\tuple{(D,\sigma),H,T}  \models \tau(i) \le \tau(k) < \tau(j)$ then $\tuple{(D,\sigma),H,T} \not \models \tr{\varphi}_{\tau(k)}$.
	By induction~\eqref{thm:kamp:ih1} we get that $(\tuple{\H,\T}, \tmf),k \not \models \varphi$, for all $i \le k < j$. 
	From all previous statements it follows $(\tuple{\H,\T}, \tmf), i \not \models  \varphi  \metricol{\release}{m}{n}\psi$: a contradiction.
	
	For the converse direction assume by contradiction that $(\tuple{\H,\T}, \tmf),i \not \models \varphi \metricol{\release}{m}{n}\psi$.
	Then, there exists $j \ge i$ such that $m \le \tau(j)-\tau(i) < n$,  $(\tuple{\H,\T}, \tmf),j \not \models \psi$ and for all $i \le k < j$, $(\tuple{\H,\T}, \tmf),k \not \models \varphi$.
	Since  $m \le \tau(j)-\tau(i) < n$ then $\tuple{(D,\sigma),H,T} \models \tau(i) \le \tau(j) \wedge \tau(i) \peq{-m} \tau(j)  \pt{n} \tau(i)$.
	From $(\tuple{\H,\T}, \tmf),j \not \models \psi$ and the induction hypothesis~\eqref{thm:kamp:ih1} we get $\tuple{(D,\sigma),H,T} \not \models \tr{\psi}_{\tau(j)}$.
	From  $(\tuple{\H,\T}, \tmf),k \not \models \varphi$, for all $i \le k < j$, and the induction hypothesis~\eqref{thm:kamp:ih1} we get $\tuple{(D,\sigma),H,T} \not \models \tr{\varphi}_{d}$ for all $d \in \lbrace\tau(k) \mid \tau(k) \in D \hbox{ and } \tau(i) \le \tau(k) < \tau(j)\rbrace$.
	By the semantics, $\tuple{(D,\sigma),H,T} \not \models \forall z \left(\tau(i) \le z < \tau(j) \to \tr{\varphi}_z\right) $.
	Therefore, $\tuple{(D,\sigma),H,T} \not \models \tr{\varphi \metricol{\release}{m}{n}\psi}_{\tau(i)}$:  a contradiction.
	
	\item The case of $\metricol{\previous}{m}{n}\psi$ is similar to the case of $\metricol{\next}{m}{n}\psi$. 
	\item The case of $\metricol{\wprevious}{m}{n}\psi$ is similar to the case of $\metricol{\wnext}{m}{n}\psi$.
	\item The case of $\varphi\metricol{\since}{m}{n}\psi$ is similar to the case of $\varphi\metricol{\until}{m}{n}\psi$. 
	\item The case of $\varphi\metricol{\trigger}{m}{n}\psi$ is similar to the case of $\varphi\metricol{\release}{m}{n}\psi$.
\end{itemize}
\end{proofof}

\end{document}